\documentclass{article}

\RequirePackage{amsthm,amsmath,amsfonts,amssymb,mathtools}
\RequirePackage[numbers]{natbib}

\usepackage{hyperref}
\hypersetup{colorlinks,citecolor=blue,urlcolor=blue}

\usepackage{graphicx}
\usepackage{setspace}
\RequirePackage[OT1]{fontenc}
\usepackage{verbatim}
\usepackage{multirow}
\usepackage{makecell}
\usepackage{tabularx}
\usepackage{tabu}
\usepackage{float}
\usepackage{stfloats}

\newcommand{\indep}{\rotatebox[origin=c]{90}{$\models$}}

\usepackage{geometry}
\onecolumn 

\usepackage[ruled,vlined]{algorithm2e}

\usepackage{enumitem}

\DeclareMathOperator*{\argmin}{arg\,min}
\DeclareMathOperator*{\argmax}{arg\,max}

\theoremstyle{plain}

\theoremstyle{remark}

\title{Reinforcement Learning in Modern Biostatistics: Constructing Optimal Adaptive Interventions}

\author{Nina Deliu$^{1,2}$, Joseph Jay Williams$^{3}$, Bibhas Chakraborty$^{4,5,6}$\\
\normalsize $^{1}$MEMOTEF Department, Sapienza University of Rome, Italy\\ \normalsize$^{2}$MRC -- Biostatistics Unit, University of Cambridge, UK\\ \normalsize$^{3}$Department of Computer Science, University of Toronto, Canada\\ 
\normalsize$^{4}$Centre for Quantitative Medicine, Duke-NUS Medical School, Singapore\\ 
\normalsize$^{5}$Department of Statistics and Data Science, National University of Singapore (NUS)\\
\normalsize$^{6}$Department of Biostatistics and Bioinformatics, Duke University, USA}

\date{}

\begin{document}
\maketitle

\begin{abstract}
In recent years, reinforcement learning (RL) has acquired a prominent position in health-related sequential decision-making problems, gaining traction as a valuable tool for delivering adaptive interventions (AIs). However, in part due to a poor synergy between the methodological and the applied communities, its real-life application is still limited and its potential is still to be realized. 
To address this gap, our work provides the first unified technical survey on RL methods, complemented with case studies, for constructing various types of AIs in healthcare. In particular, using the common methodological umbrella of RL, we bridge two seemingly different AI domains, dynamic treatment regimes and just-in-time adaptive interventions in mobile health, highlighting similarities and differences between them and discussing the implications of using RL. Open problems and considerations for future research directions are outlined. Finally, we leverage our experience in designing case studies in both areas to showcase the significant collaborative opportunities between statistical, RL, and healthcare researchers in advancing AIs.

\end{abstract}

\section{Introduction}\label{Sec: Intro}
In the era of big data and digital innovation, healthcare is going through a rapid and dramatic change process, transitioning from \textit{one-size-fits-all} standards to the tailored approach of \textit{precision} or \textit{personalized medicine}~\citep{kosorok_precision_2019}. Under this framework, the ``individual variability in genes, environment, and lifestyle for each person'' is taken into account in an effort to improve the ways we ``anticipate, prevent, diagnose, and treat'' a particular disease in a particular patient~\citep{collins_new_2015}. 
This paradigm encompasses a broad range of scientific domains, ranging from genomics to advanced analytics and causal inference, all in support of a data-driven, yet patient-centric, approach for delivering personalized care.

One of the key methodological lines of research within the domain of personalized medicine is the development of \textit{adaptive interventions} (AIs)~\citep{almirall_introduction_2014, collins_conceptual_2004}. The fundamental goal of AIs is to operationalize sequential decision-making by tailoring interventions to individuals, so as to offer guidance on how to adapt them to an individual's changing status and needs. In clinical practice, a typical situation is represented by a clinician who needs to use a set of treatment rules (i.e., a treatment regime) that recommend how to assign treatments or doses to patients based on their individual characteristics. These characteristics can include both baseline information (e.g., demographic data or pretreatment clinical conditions) and evolving health status (e.g., responses to previous treatments). For example, for patients who do not improve on the first-line treatment over a prespecified period, the clinician may plan to increase the dose, according to a dose-response relationship, or change treatment in the case of a sensitive or drug-resistant patient. Due to changes in their health status, such a treatment regime is therefore \textit{dynamic within} a person. To the patient, this sequence of treatments seems like standard treatment; to the clinician, it represents a series of prespecified decisions to make according to the patient's evolving history; and to the statistician, it constitutes an AI, alternatively known as \textit{dynamic treatment regime} or \textit{regimen} (DTR)~\citep{murphy_optimal_2003, lavori_dynamic_2004, chakraborty_dynamic_2014}. The distinctive feature of AIs is their data-driven, adaptive approach guided by and oriented toward individual data. Clearly, an ambitious goal in AIs, or more specifically in DTRs, is how to construct the \textit{optimal} DTRs, e.g., treatment regimes that result in an optimal mean response or outcome. Such a question has a long history in statistics and its study will occupy a central role in this work.


The traditional way of offering AIs to a patient mostly relies on rules created by experts, based on factors such as domain theory and empirical experience with similar patients. However, the recent advances and the widespread application of artificial intelligence and machine learning (ML) techniques (see e.g.,~\cite{deo_machine_2015,rajkomar_machine_2019,oyebode_machine_2022}), have shed light on their ability to enable clinicians to quickly, efficiently, and accurately identify the most appropriate course of action for their patients.

ML represents a hotspot in artificial intelligence, and health systems have recently tapped into its expanding potential to complement classical statistical tools and support clinical decision-making. There is no clear line between ML models and traditional statistical models~\citep{beam_big_2018}. Yet, it is widely acknowledged that sophisticated ML models such as deep learning models~\citep{goodfellow2016deep} excel in learning, and automatically improving through experience, from the high-dimensional and heterogeneous data generated from modern clinical care. By matching a patient's characteristics to a computerized clinical knowledge base, such algorithms can suggest assessments or recommendations tailored to that patient's characteristics, even in very complex settings. Despite the potential to revolutionize decision-making, the success of ML in healthcare strongly depends on the efforts of the theoretical and methodological communities to unravel and elucidate their intrinsic mechanism and processes (often criticized for operating in a `black box'). In turn, this may foster broader acceptance among clinicians and patients, thereby fostering greater confidence in the integration of ML technologies into clinical practice. 

Among the existing ML paradigms~\citep{bishop_pattern_2006, mohri_foundations_2018}, \textit{reinforcement learning} (RL)~\citep{sutton18_rl,bertsekas2019reinforcement, sugiyama2015statistical} offers a natural framework for the sequential decision-making problem encountered in AIs. 
In classical RL, a \textit{learning agent} has to decide which of one or more \textit{actions} to take when interacting with an \textit{unknown environment}. Based on the feedback or \textit{reward} received from the environment for the selected action(s), the agent learns how best to act to maximize the cumulative reward over time. This is done by \textit{trial-and-error}, that is, by observing and inferring from the environment after actions are taken. The RL framework is abstract and yet flexible enough to accommodate a variety of domains where the problem has a sequential nature~\citep{chakraborty_statistical_2013, gottesman_guidelines_2019}; it does so by specifically characterizing the environment's (or domain's) dynamics. In AIs, RL can be applied by regarding the alternative interventions as the actions to be chosen and the outcome of the intervention (e.g., patient's response) as the reward; patient's time-varying context and status represent the environment.

Within biostatistics, RL was first introduced as a data analysis tool to discover optimal DTRs in a variety of health domains including cancer~\citep{zhao_reinforcement_2009, goldberg_q-learning_2012}, weight loss management~\citep{forman2019can, pfammatter_smart_2019}, substance use~\citep{murphy_developing_2007, chakraborty_statistical_2013}, mental health~\citep{pike2022reinforcement}, and so on. More recently, there seems to be an unprecedented interest in the application of RL in the rapidly expanding \textit{mobile health} (mHealth)~\citep{istepanian2007m, kumar2013mobile, rehg2017mobile} domain. The mHealth area refers to the use of mobile or wearable technologies to promote healthy behavior changes in both clinical and nonclinical populations. A high-level goal in mHealth is to deliver efficacious \textit{just-in-time adaptive interventions} (JITAIs)~\citep{nahum2018just} in response to the \textit{in-the-moment} changes in an individual's internal state (e.g., health) and contextual state (e.g. location)~\citep{rehg2017mobile}. The challenge in JITAIs is thus to provide `the right individual with the right intervention', as well as `the right intervention at the right time'.  Notably, despite the relatively recent development of JITAIs compared to DTRs, research interest in both methodology and applications has substantially skewed toward JITAIs; we refer to Figure 1 in Supplementary Material A for a quantification of the volume of the literature. 
Given the increasing number of mHealth studies and in tandem the ongoing interest among statisticians in DTRs, integrating these two areas is a worthy objective. In the current article, we combine our methodological background with our experience in designing case studies in the above two areas to extensively review the state of the art of RL in AIs. To the best of our knowledge, this represents the first comprehensive survey of RL methods for developing DTRs as well as JITAIs in mHealth, informed by our experience with the challenges and successes of real-world applications. It complements and adds to the extensively surveyed DTR literature (see e.g.,~\cite{chakraborty_statistical_2013, chakraborty_dynamic_2014, tsiatis_dynamic_2021}), which we place together with JITAIs under the same AI umbrella.

We believe that there is ample scope for important practical advances in these areas, and with this survey we aim to make it easier for theoretical and methodological researchers to join forces to assist healthcare discoveries by developing the next generation of methods for AIs in healthcare. 
We finally emphasize that we focus on healthcare and biostatistics due to the central role statisticians play there traditionally. Notwithstanding, the concepts we review for AIs extend far beyond: to education~\citep{nahum-shani_introduction_2019}, policy making~\cite{kasy_adaptive_2021}, and other domains such as population research, where RL has recently been contextualized~\cite{deliu_reinforcement_2023}.

The remainder of this work is structured as follows. In Section~\ref{Sec: RLforAIs}, we formally characterize the problem of AIs, providing a common framework for applications to DTRs and JITAIs, and explaining their similarities and differences. We then formalize the RL paradigm and its subclasses, relating it to the problems at hand, and assimilating the different existing notations and terminologies into a coherent framework (Section~\ref{Sec: RL_framework}). This provides a foundation for conducting research more easily in both methodological and applied aspects of AIs, enhancing communication and synergy between them. Section~\ref{Sec: RLmethods} offers a review of RL methods for developing AIs, expanding on DTRs with both finite- and indefinite-time horizons and JITAIs for mHealth. Insights on current methodological differences, along with their drivers, are discussed in Section \ref{sec: divergence_DTR_JITAI} and considerations for future research are provided in Section \ref{sec: future_durections}. Section~\ref{Sec: AIs_RealLife} grounds Section~\ref{Sec: RLmethods} by illustrating the development and application of the presented methodology to two case studies. Section~\ref{Sec: Discussion} concludes with some final remarks.

\section{Adaptive Interventions in Healthcare} \label{Sec: RLforAIs}

Adaptive interventions offer a vehicle to operationalize a sequential decision-making process over the course of a program or a condition, with the aim of optimizing individual outcomes. Technically speaking, AIs are defined via explicit sequences of decision rules that prespecify how the type, intensity, and delivery of intervention options should be adjusted over time in response to individual progresses~\citep{almirall_introduction_2014, nahum2018just}.  The prespecified nature of AIs increases their replicability in research and enhances the assessment of their effectiveness~\citep{nahum-shani_introduction_2019}. 

\begin{figure*}[htb]
    \centering
    \includegraphics[width=.9\linewidth]{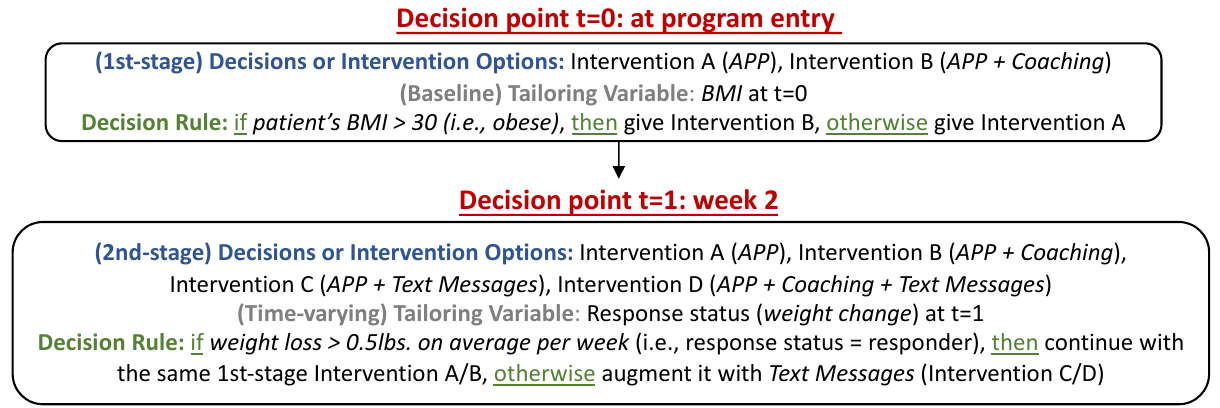}
    \caption{Simplified schematic of a two-stage AI and its key components, inspired by the weight loss management study in~\citep{pfammatter_smart_2019}}
    \label{fig: ai-example}
\end{figure*}

The existing frameworks for formalizing AIs~\citep{collins_conceptual_2004, almirall_introduction_2014} are based primarily on four key components:
\begin{enumerate}[label = (\roman*)]
    \item The \textbf{decision points}, specifying the time points or time intervals at which a decision concerning intervention has been or has to be made; here, we assume a finite or countable number of times $t \in \mathbb{N} = \{0, 1,\dots\}$;
    \item The \textbf{decisions or intervention options} at each time $t$, that may correspond to different types, dosages (duration, frequency or amount~\citep{voils_informing_2012}), or delivery modes, as well as various tactical options (e.g., augment, switch, maintain); we denote them by $A_t \in \mathcal{A}_t$, where $\mathcal{A}_t$ is the decision or action space, generally discrete, at time $t$;
    \item The \textbf{tailoring variable(s)} at each time $t$, say $X_t \in \mathcal{X}_t$, with $\mathcal{X}_t \subseteq \mathbb{R}^n, n \geq 1$, capturing individuals' baseline and time-varying information for personalizing decision-making;
    \item The \textbf{decision rules}, $\boldsymbol{d} = \{d_t\}_{t \geq 0}$, where, at each time $t$, $d_t$ links the tailoring variable(s) $X_t$ and potentially any other previous information deemed important to a specific decision or intervention $A_t \in \mathcal{A}_t$.
\end{enumerate}
A common illustrative way to describe an AI is through schematics such as the one shown in Figure~\ref{fig: ai-example}, where the ``if-then'' statements clarify how the decision rule prespecifies the intervention options under various conditions.

Since an AI adaptation is aimed at optimizing individual outcomes, these play an essential role when defining an AI's components~\citep{nahum-shani_introduction_2019}. In particular, we can distinguish between two types of individual outcomes: 
\begin{enumerate}
    \item[(v)] The \textbf{intermediate} or \textbf{proximal outcome(s)}, say $\{Y_t\}_{t>0}$, with $Y_t \in \mathcal{Y}_t$, that is, easily observable short-term outcome(s), expected to influence a longer-term outcome according to some mediation theory~\citep{mackinnon_mediation_2007};
    \item[(vi)] The \textbf{final} or \textbf{distal outcome}, representing the long-term outcome of interest and the ultimate goal of the AI. To distinguish it from the intermediate stage-related outcomes $\{Y_t\}_{t>0}$, which may have a different meaning and nature, we denote it by $\overset{\infty}{Y}$. 
\end{enumerate}
Different AI problems would target different types of outcomes. For example, in some contexts, there may only be a distal (end-of-study) outcome $\overset{\infty}{Y}$ instead of multiple intermediate outcomes (see e.g.,~\cite{pelham_effects_2002}): in that case, we will use the convention $Y_{T+1} \doteq \overset{\infty}{Y}$, with $T$ being the study's last decision point or problem horizon. In other cases, only the intermediate outcomes will define the AI problem, while the final outcome will have no formal role in the decision-making problem. We also note that proximal outcomes can also be used as tailoring variables to guide later-stage decisions. In Figure~\ref{fig: ai-example}, for example, the response status at time $t=1$ represents both the proximal outcome targeted by the intervention at the decision point $t=0$ and the tailoring variable at the decision point $t=1$.

Development of AIs is based on the selection and integration of the aforementioned six components, taking into account their relationship. Ideally, this is informed and guided by domain theories, practical considerations, empirical evidence, or some combinations thereof. Determining optimized decision rules typically involves more sophisticated data-driven statistical and ML tools, with RL recognized as a current state-of-the-art tool.

The term AI is interchangeably used with \textit{adaptive treatment strategy}~\citep{murphy_experimental_2005, murphy2007customizing}, \textit{treatment policy}~\citep{lunceford_estimation_2002, wahed_semiparametric_2006, dawson_efficient_2012}, and 
\textit{dynamic treatment regime} or \textit{regimen}~\citep{murphy_optimal_2003, lavori_dynamic_2004, chakraborty_statistical_2013, laber_dynamic_2014}, among others. However, given its more generic nature, we use the term AI to refer to a general framework for personalizing interventions sequentially based on an individual's time-varying characteristics. This broader definition embraces a considerate number of applications, including non-healthcare (e.g., education~\citep{nahum-shani_introduction_2019}) and the two healthcare domains of DTRs and JITAIs, which we cover below. 

\subsection{Dynamic treatment regimes} In medical research, DTRs define a sequence of treatment rules tailored to each individual patient based on their baseline and time-varying (dynamic) state. 
Traditionally, treatment assignment is based on \textit{single-stage} decision-making. Specifically, one observes a set of baseline or pretreatment information $X_0 \in \mathcal{X}_0$, based on which a treatment $A_0  \in \mathcal{A}_0$ is selected. The treatment rule, say $d_0$, is a mapping from $\mathcal{X}_0$ to $\mathcal{A}_0$. If more stages are involved, at each stage $t$, the treatment rule $d_t$ is again an (independent) mapping from the stage-$t$ information set $\mathcal{X}_t$ to a stage-$t$ action space $\mathcal{A}_t$. Unlike single-stage protocols, where $\boldsymbol{d}_t = \{d_\tau\!: \mathcal{X}_\tau \to \mathcal{A}_\tau\}_{\tau = 0,\dots, t}$, DTRs explicitly incorporate the heterogeneity in treatment effect among individuals and \textit{across time} within an individual, and regards $\boldsymbol{d}_t$ as a (nested) \textit{multistage regime} with each $d_\tau$, $\tau = 0,\dots,t$, depending on the individual evolving history of covariates, treatments, and outcomes up to time $\tau$; that is, $\boldsymbol{d}_t = \{d_\tau\!: \mathcal{H}_\tau \to \mathcal{A}_\tau\}_{\tau = 0,\dots, t}$, with $\mathcal{H}_\tau \doteq \mathcal{X}_0\times\mathcal{A}_0 \times \mathcal{Y}_1 \times \dots \times\mathcal{A}_{\tau-1} \times \mathcal{Y}_{\tau} \times \mathcal{X}_\tau$. As such, it provides an attractive framework for personalized treatments in longitudinal settings. Beyond personalization, DTRs can identify and evaluate delayed effects, i.e., effects that do not occur immediately after treatment but may affect a person or their disease later in time. It should also be noted that, by treating only those who show a need for treatment, DTRs hold the promise of reducing noncompliance due to overtreatment or undertreatment~\citep{lavori_design_2000}. At the same time, they are attractive to public policy makers, allowing a better allocation of public and private funds~\citep{murphy_optimal_2003}. 

\begin{figure*}[htb]
    \centering
    \includegraphics[width=1\linewidth]{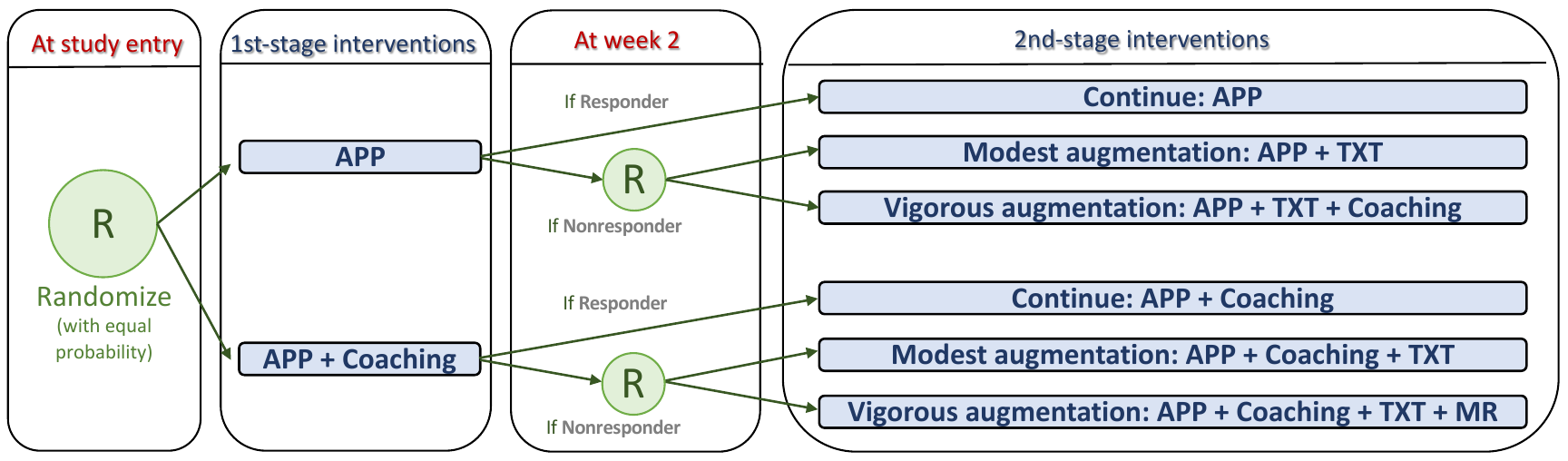}
    \caption{Schematic of the weight loss SMART in~\citep{pfammatter_smart_2019}}
    \label{fig: smart-wl}
\end{figure*}

For developing DTRs, data sources include both longitudinal observational data, such as electronic health records (EHRs), and randomized studies, such as randomized-controlled trials (RCTs) and \textit{sequential multiple assignment randomized trials} (SMARTs)~\citep{lavori_design_2000, murphy_experimental_2005}. Although observational sources are much more common, SMARTs represent the current gold standard~\citep{lei_smart_2012}.
A SMART is characterized by multiple stages of treatment, typically ranging from two to four, where each stage corresponds to one of the critical decision points. 
A concrete example is provided in~Figure~\ref{fig: smart-wl}, which illustrates the first two stages of the weight loss management study in~\citep{pfammatter_smart_2019}.
At study entry, all individuals are uniformly randomized to one of two first-line interventions: mobile app (APP) or APP + Coaching. Participants are assessed at weeks 2, 4, 8, and those `responding' to their initial treatment (i.e., losing at least 0.5 lbs. on average per week) continue receiving the same treatment. As soon as an individual is classified as a `nonresponder', they are re-randomized to one of two augmentation tactics: modest augmentation (supportive text message; TXT) or vigorous augmentation (TXT + Coaching, or TXT + meal replacement (MR), depending on the first-stage treatment). Rerandomization occurs only once per participant, with the newly assigned treatment continuing through the end of the study. 
Because different intervention options are considered for responders (continue) and nonresponders (modest or vigorous augmentation), the response status is embedded as a tailoring variable. Such multistage restricted randomization generates several DTRs \textit{embedded} in the SMART; we refer to~\cite{chakraborty_statistical_2013} for details on embedded regimes.

\subsection{Just-in-time adaptive interventions in mHealth} 
The ubiquitous use of mobile technologies has facilitated the development of a new area of health promotion in both clinical and nonclinical populations, known as mHealth~\citep{istepanian2007m}. 
A key objective in mHealth is to deliver efficacious real-time AIs in response to rapid changes in individual circumstances, while avoiding overtreatment and its consequences on user engagement (e.g., low adherence to recommendations or discontinued usage of the mobile device). This specialized AI is termed just-in-time adaptive intervention (JITAI)~\citep{nahum2018just}. JITAIs are nowadays gaining an increased popularity across various behavioral domains, spanning from physical activity \citep{hardeman2019systematic,figueroa_daily_2022} and weight management \citep{pfammatter_smart_2019} to addictive disorders \citep{goldstein2017return, garnett2019development,naughton2017delivering} and mental health \citep{, kumar_using_2024}. Moreover, there has also been recent interest in leveraging JITAIs to enhance public health on a broader scale~\citep{liu_microrandomized_2023}.

\begin{figure*}[htb]
   \centering
   \includegraphics[width=1\linewidth]{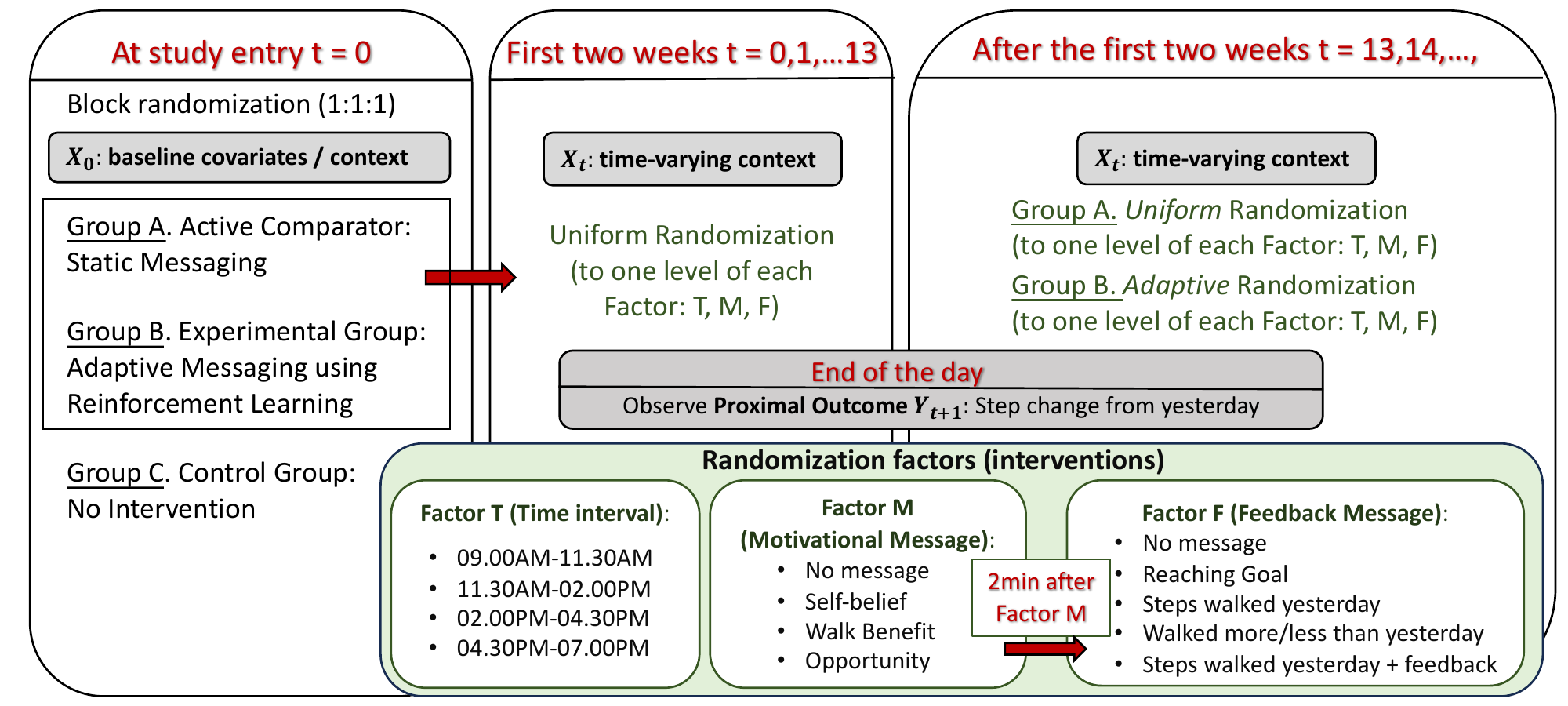}
   \caption{Schematic of the DIAMANTE micro-randomized trial~\citep{aguilera2020mhealth}} \label{fig: mrt-diamante}
\end{figure*}

In mHealth, JITAIs refer to a sequence of decision rules that use continuously collected data through mobile technologies (e.g., wearable devices, accelerometers, or smartphones) to adapt intervention components in real time in order to support behavior change and to promote health. The peculiarity of JITAIs is that they deliver interventions according to the user’s in-the-moment context or needs, e.g., time, location, or current activity, including considerations of whether and when the intervention is needed. Compared to DTRs, JITAIs are more flexible in terms of location and timing of interventions delivery. In fact, while the adaptation and delivery of a DTR usually take place at a pre-defined clinical appointment and under the direct guidance of a clinician, JITAIs often adapt and assign interventions as dictated by the mobile system or individual users, while they go about their daily lives in their natural environments. For this reason, unless otherwise designed, the time interval between decision points can vary significantly between and within subjects, dictated by randomness in individual needs and engagement with the mHealth device. Furthermore, unlike DTRs, the number of decision points in JITAIs can be hundreds or even thousands, and the intervention can be delivered each minute, hour, or day (as in the case of the \textit{DIAMANTE} study, which will be shortly discussed and illustrated in Figure~\ref{fig: mrt-diamante}). 

In JITAIs, the time between decision points is often too short to capture the (distal) clinical outcome of interest, and they rely on a weak surrogate, i.e., the proximal outcome. Unlike DTRs--which target the distal outcome and may or may not have an intermediate (proximal) outcome--in JITAIs, proximal outcomes represent the direct and in-the-moment target of the intervention. The distal outcome is expected to improve only based on domain knowledge about its relationship with the proximal outcome, but is not formally included in the optimization problem. We refer to Table~\ref{tab: dtr_ai_diff2} for a hand-to-hand comparison between JITAIs and DTRs under the AI framework. 

\begin{table*}
\centering
\footnotesize
    \caption{Differences between DTRs and JITAIs in terms of the key characteristics defining an AI}
    \begin{tabu} to \textwidth { X[.9] | X[1.7] | X[1.5] }
        \multicolumn{3}{c}{}\\
        \multicolumn{1}{c}{\textbf{Characteristic}
        }
        &\multicolumn{2}{c}{\textbf{Type of Adaptive Intervention}}\\
        \hline \\
        & \textbf{DTRs} & \textbf{JITAIs in mHealth}\\
        \hline
        Data sources & RCTs, SMARTs, longitudinal observational data including EHRs, dynamical systems models & MRTs, RCTs, factorial designs, single-case experimental designs\\
        \hline
        \multicolumn{3}{c}{}\\
        \multicolumn{3}{l}{\textbf{AI component: (i) decision points $t \in \mathbb{N}$}}\\
        \hline
        Number of decision points & In SMARTs, generally small (e.g., two to four); in EHRs (defined over indefinite horizons) an increased number is seen  & Generally very large (hundreds or even thousands for each single unit)\\
        \hline
        Alignment of decision points across subjects & In SMARTs, these are expected to occur at some regular and fixed time points; in EHRs, these can reflect different protocols and a higher variability between individuals, with less regular patterns. & Since decision points reflect user's specific needs and availability, in JITAIs, these are often random (as requested by the user).\\
        \hline 
        Distance between decision points & Sufficiently long according to the expected time to capture a potential effect (including a delayed effect) of the intervention on the primary outcome of interest or a strong intermediate surrogate & Quite short according to the expected ``in-the-moment'' effect of the intervention on the proximal outcome (e.g., every few minutes, hours or daily)\\
        \hline
        \multicolumn{3}{c}{}\\
        \multicolumn{3}{l}{\textbf{AI component: (ii) decisions or intervention options $A_t \in \mathcal{A}_t, t \in \mathbb{N}$}}\\
        \hline
        Type of intervention & Mostly drugs or behavioral interventions & Generally behavioral interventions (e.g., motivational/feedback messages, coaching, reminders) with few exceptions (e.g., insulin adjustments)\\
        \hline
        Intervention delivery & Assigned by the care provider during an appointment or through digital devices & Assigned through digital/mobile devices according to an automatic algorithm or/and under care provider's guidance\\
        \hline 
        \multicolumn{3}{c}{}\\
        \multicolumn{3}{l}{\textbf{AI component: (iii) tailoring variable(s) $X_t \in \mathcal{X}_t \subseteq \mathbb{R}^n, n \geq 1, t \in \mathbb{N}$}}\\
        \hline
        Type of tailoring variable & Can include the full or partial history of baseline and time-varying patients' information. An external context can also be considered, but it has secondary relevance. & Current users' information, and any type of variable related to their momentary context (e.g., availability, weather), which plays a major role and can be very granular\\
        \hline
        \multicolumn{3}{c}{}\\
        \multicolumn{3}{l}{\textbf{AI component: (iv) decision rules $\boldsymbol{d} = \{d_t\}_{t \in \mathbb{N}}$}}\\
        \hline
        Main strategy to optimize decision rules &  \begin{itemize}[leftmargin=*, noitemsep, topsep=0pt] \item Offline methods for finite-horizon decision problems, with some exceptions (e.g., for EHRs-based DTRs an indefinite horizon may be considered) \item While finite-horizon problems in general account for the full individual history over time, indefinite horizon problems assume a Markov structure. \end{itemize} & \begin{itemize}[leftmargin=*, noitemsep, topsep=0pt] \item Online methods over indefinite-time horizons \item Considering the expected ``in-the-moment'' effect of the intervention, typically, only the current or last observed information is accounted for, with a predominant use of Markov, partially observed Markov, or simpler structures \end{itemize}\\
        \hline
        \multicolumn{3}{c}{}\\
        \multicolumn{3}{l}{\textbf{AI component: (v)-(vi) outcomes}}\\
        \hline
        Proximal outcome\\ $Y_t \in \mathcal{Y}_t \in \mathbb{R}, t=1,2,\dots$ &  \begin{itemize}[leftmargin=*, noitemsep, topsep=0pt] \item \textit{Optional} short-term outcomes expected to impact the distal (long-term) outcome \item While not being the primary target of the intervention, they may be part of the adaptation/optimization process. \end{itemize} & \begin{itemize}[leftmargin=*, noitemsep, topsep=0pt] \item Short-term outcomes directly targeted by the intervention and expected to mediate the effect on the distal outcome \item They guide the definition of just-in-time in the context of the identified problem, as well as the formulation of the adaptation strategy. \end{itemize}\\  
        \hline
        Distal outcome $\overset{\infty}{Y}$ & \begin{itemize}[leftmargin=*, noitemsep,topsep=0pt] \item The outcome directly and formally targeted by the intervention \item The primary criterion that guides the adaptation/optimization of the DTR, although intermediate outcomes are often part of the optimization \end{itemize} & \begin{itemize}[leftmargin=*, noitemsep,topsep=0pt] \item Long-term goal of a JITAI, expected to be influenced by an intervention through the mediating role of proximal outcomes (domain knowledge) \item Typically, they do not guide the adaptation/optimization of the learning strategy. \end{itemize} \\
        \hline
    \end{tabu}
    \label{tab: dtr_ai_diff2}
\end{table*}

Typical experimental designs for building JITAIs are represented by \textit{factorial experiments}~\citep{collins_comparison_2009}, or most notably, \textit{micro-randomized trials} (MRTs)~\citep{klasnja2015microrandomized}. In MRTs, individuals are randomized hundreds or thousands of times over the course of the study, and in a typical multicomponent intervention study, the multiple components can be randomized concurrently, making micro-randomization a form of a sequential full factorial design. The goal of these trials is to optimize mHealth interventions during the trial while offering a basis to assess the causal effects of each intervention component and to evaluate whether the intervention effects vary with time and/or with the individual contexts. 

To better understand the characteristics and value of MRTs, let us now consider the \textit{DIAMANTE} study for promoting physical activity, illustrated in Figure~\ref{fig: mrt-diamante}.
In this study, the intervention components 
include whether or not to send a text message, which type of message to deliver, and at which time. The latter thus has a central role in this type of AI, as it defines the intervention set. The proximal outcome is the change in the number of steps a participant walked today from yesterday; and the context is given by a set of variables such as health information and study day.
To assess the effectiveness of the optimized JITAI, 
users are assigned to different study groups: (A) a static (nonoptimized) group, (B) an adaptive group based on RL, and (C) a control group. In the two intervention groups, users are randomized every day to receive a combination of categories of the different intervention components, delivered within different time intervals. The adaptive RL-based optimized group will be briefly discussed in Section~\ref{Sec: AIs_RealLife}, after introducing the RL framework.

\section{The Reinforcement Learning Framework} \label{Sec: RL_framework}
Generally speaking, RL is an area of ML concerned with determining optimal action selection policies in sequential decision-making problems \citep{sutton18_rl, bertsekas2019reinforcement}. This framework is based on repeated interactions between a \textit{decision maker} or \textit{learning agent} and the \textit{environment} it wants to learn about, to take better \text{decisions} or \textit{actions}. Before characterizing this process and formalizing the RL problem(s), it is paramount to set out clearly the fundamental prerequisites that enable RL to solve decision-making problems such as in AIs with rigor. 

\subsection{A preliminary note: causal inference and RL} \label{sec: causal_inference}

While in this work our focus is primarily on RL, we note that this is neither a necessary nor generally a sufficient solution for building valid AIs. As we will mention in passing in Section~\ref{Sec: DTRs_excursus}, a variety of other traditional statistical approaches, mostly confined to the causal inference literature, exist and have a substantial relevance in the field. In fact, for developing AIs, one needs to assess the causal relationship between interventions and outcomes, thus, requiring an adequate framework for causality. 

Causal inference provides a set of tools and principles that allows one to combine data and causal assumptions about the environment to reason with questions of counterfactual nature. Through considerations on study designs, estimation strategies, and certain fundamental assumptions (e.g., no unmeasured confounders), it provides the building blocks that enable researchers to draw causal conclusions based on the observed data.
On a different tangent, RL is concerned with efficiently finding a policy that optimizes an objective function (e.g., the expected cumulative reward) in interactive and uncertain environments. In practice, despite being causal by nature--any system looking to advise on interventions in some way quantifies their effects--the classical RL does not conduct causal inference. We can think of at least two reasons. First, RL practitioners often consider problems in which the data are unconfounded (e.g., robotics), because these are collected through direct interactions with a relatively well-understood environment, governed by physical laws, and actions are taken by the learning agent depending only on the data available (experimental data). To illustrate, this is the case of current JITAI practices. Second, and most importantly, the fundamental problem of RL, rather than dealing with causal effects estimation, is oriented toward causal-decision making. We note that the two are not the same, and counterintuitively, accurate estimation is not essential for accurate decision-making~\citep{fernandez-loria_causal_2022}.
While these two areas have evolved independently over different aspects of the same building block and with no interaction between them, disciplines such as AIs can be developed only under an integrated framework that permits causal conclusions. 

Our attention in the current work is devoted to RL rather than causal inference, and we point the readers to the seminal works of Neyman and Rubin~\cite{neyman1923application,rubin1974estimating} for the potential outcomes framework, and to Pearl~\cite{Pearl09a} for the causal graphical model perspective. For a comprehensive treatment of both, we refer to~\citep{hernan_causal_2023}. Furthermore, recent attempts in the ML community have worked toward a unified framework called causal RL, which embeds the causal graphical approach within sample efficient RL algorithms~\citep{zhang_designing_2020}. 
For simplicity of exposition, in this work, we assume that the main assumptions of causal inference (see e.g.,~\cite{chakraborty_dynamic_2014}) hold, and that the conditional distributions of the observed data are the same as the conditional distributions of the potential outcomes, given the assigned treatment. It follows that RL can operate in a simplified causal inference problem (in which actions are unconfounded), and that optimal AIs may be obtained using the observed data.


\subsection{Formalization of the general RL problem} \label{Sec: full-RL}
Consider a discrete time space indexed by $t \in \mathbb{N}= \{0, 1,\dots, \}$. In RL, at each decision time point or simply time $t$, an agent faces a decision-making problem in an unknown environment. After receiving some representation of the environment's \textit{state} or \textit{context}, say $X_t \in \mathcal{X}_t$, it selects an \textit{action}, denoted by $A_t$, from a set of admissible actions $\mathcal{A}_t$. As a result, one step later, the environment responds to the agent's action by making a transition into a new state $X_{t+1} \in \mathcal{X}_{t+1}$ and (typically) providing a numerical \textit{reward} $Y_{t+1} \in \mathcal{Y}_{t+1} \subset \mathbb{R}$. By repeating this process over time, the result is a trajectory 
of states visited, actions pursued, and rewards received. 
In a medical context, this trajectory can be viewed as the individual \textit{history} (of covariates, treatments, and responses to treatments) of a patient over time. Note that in some settings there may be only one terminal reward (or a final outcome, e.g., overall survival or school performance at the end of the study~\citep{pelham_effects_2002}); in this case, rewards at all previous time points are taken to be $0$. In other settings (e.g., multi-armed bandits; Section \ref{sec:MAB}), states may be ignored, thus leading to a trajectory of actions and rewards only.

Define $\mathbf{X_t} \doteq \left(X_0, \dots, X_t \right)$, $\mathbf{A_t} \doteq \left(A_0, \dots, A_t \right)$, $\mathbf{Y_{t+1}} \doteq \left(Y_1, \dots, Y_{t+1} \right)$, and similarly $\mathbf{x_t}$, $\mathbf{a_t}$ and $\mathbf{y_{t+1}}$, where the upper- and lower-case letters denote random variables and their particular realizations, respectively. Define \textit{history} $\mathbf{H_t}$ as all the information available at time $t$ prior to decision $A_{t}$, i.e., $\mathbf{H_t}  \doteq (\mathbf{A_{t-1}}, \mathbf{X_t}, \mathbf{Y_t})$; similarly $\mathbf{h_t}$. The history $\mathbf{H_t}$ at time $t$ belongs to the product set $\boldsymbol{\mathcal{H}_t} = \mathcal{X}_0 \times \prod_{\tau = 1}^{t}{\mathcal{X}_\tau \times \mathcal{A}_{\tau-1} \times \mathcal{Y}_\tau}$. Note that, by definition, $\mathbf{H_0} = X_0$. 
We assume that each longitudinal history is sampled independently according to a distribution $P_{\boldsymbol{\pi}}^{\text{Full-RL}}$ (with the superscript clarified later in Section \ref{sec: MDP-MAB}), 
given by
\begin{align} \label{eq: P_pi}
P_{\boldsymbol{\pi}}^{\text{Full-RL}} \doteq p_0(x_0) \prod_{t \geq 0} \pi_{t}(a_{t} | \mathbf{h_{t}})
p_{t+1}(x_{t+1}, y_{t+1}|\mathbf{h_{t}}, a_{t}),
\end{align} 
where:
\begin{itemize}[leftmargin=*]
    \item $p_0$ is the probability distribution of the initial state $X_0$.

    \item $\boldsymbol{\pi} \doteq \{\pi_t\}_{t \geq 0}$ represents the \textit{exploration policy} that determines the sequence of actions generated throughout the decision-making process. More specifically, $\pi_t$ maps histories of length $t$, $\mathbf{h_t}$, to a probability distribution over the action space $\mathcal{A}_t$, i.e. $\pi_t( \cdot |\mathbf{h_t})$. The conditioning symbol `$|$' in $\pi_t( \cdot |\mathbf{h_t})$ reminds us that the exploration policy defines a probability distribution over $\mathcal{A}_t$ for each $\mathbf{h_t} \in \boldsymbol{\mathcal{H}_t}$. 
    Sometimes, $A_t$ is uniquely determined by the history $\mathbf{H_t}$, therefore, the policy is simply a function of the form $\pi_t(\mathbf{h_t}) = a_t$.  We call it \textit{deterministic policy}, in contrast with \textit{stochastic policies} that determine actions probabilistically.

    \item $\{p_t\}_{t \geq 1}$ are the unknown \textit{transition probability distributions} and they completely characterize the dynamics of the environment. At each time $t \in \mathbb{N}$, the transition probability $p_t$ assigns to each trajectory $(\mathbf{x_{t-1}},\mathbf{a_{t-1}}, \mathbf{y_{t-1}}) = (\mathbf{h_{t-1}},a_{t-1})$ at time $t-1$ a probability measure over $\mathcal{X}_t \times \mathcal{Y}_t$, 
    i.e., $p_t(\cdot, \cdot|\mathbf{h_{t-1}}, a_{t-1})$. 
\end{itemize}

At each time $t$, the transition probability distribution $p_{t+1}(x_{t+1}, y_{t+1}|\mathbf{h_{t}}, a_{t})$ gives rise to: 
\begin{itemize}
\item[(i)] $p_{t+1}(x_{t+1}|\mathbf{h_{t}}, a_{t})$, the \textit{state-transition probability distribution}, representing the probability of moving to state $x_{t+1}$ having observed history $\mathbf{h_{t}}$ and taking action $a_{t}$;
\item[(ii)] $p_{t+1}(y_{t+1}|\mathbf{h_{t}},a_{t},x_{t+1})$, the \textit{immediate reward distribution}, specifying the reward $Y_{t+1}$ after transitioning to $x_{t+1}$ with action $a_t$. 
\end{itemize}
Generally, in DTRs, the immediate reward $Y_{t+1}$ is conceptualized as a known function of the history $\mathbf{H_t}$, the current selected action $A_t$ and the new state $X_{t+1}$; 
that is, conditional on $\mathbf{H_t}$, the reward function is deterministic and $Y_{t+1}$ is uniquely determined. To give a concrete example, one can think of a dose-finding trial, where the level of toxicity is one of the state variables, among others. In this setting, at each time $t$, the immediate reward $Y_{t+1}$ of a patient with history $\mathbf{H_t}$ and administered dose $A_t$ could be defined as a binary variable assuming value $-1$ if the observed toxicity level ($X_{t+1}$) is higher than a certain pre-specified threshold, and $0$ otherwise. 

The cumulative sum (often time-discounted) of immediate rewards is termed \textit{return}, say $\mathbf{R_t}$, and is given by
\begin{align} \label{eq: return}
\mathbf{R_t} \doteq Y_{t+1} + \gamma Y_{t+2} + \gamma^2 Y_{t+3} + \dots = \sum_{\tau \geq t} {\gamma^{\tau-t} Y_{\tau+1}},
\end{align}
for $t \in \mathbb{N}$. The \textit{discount rate} $\gamma \in [0, 1]$ determines the current value of future rewards: a reward received $\tau$ time steps in the future is worth only $\gamma^{\tau}$ times what it would be worth if it were received immediately. If $\gamma < 1$, the potential infinite sum in Eq.~\eqref{eq: return} has a finite value as long as the reward sequence $\{Y_{\tau+1} \}_{\tau \geq t}$ is bounded. If $\gamma = 0$, the agent is \textit{myopic} in being concerned only with maximizing the immediate reward, i.e. $\mathbf{R_t} = Y_{t+1}$; this is often the case of the multi-armed bandit framework (see Section \ref{sec:MAB}). If $\gamma = 1$, the return is \textit{undiscounted} and it is well defined (finite) as long as the time horizon is finite, i.e., $t \in [0, T]$, with $T < \infty$ \citep{sutton18_rl}. If $T$ is fixed and known in advance, e.g., in clinical trials, the agent faces a \textit{finite-horizon} problem; if $T$ is not prespecified and can be arbitrarily large (the typical case of EHRs), but finite, we call it an \textit{indefinite-horizon} problem; finally we use the term \textit{infinite-horizon} problem when $T = \infty$. In this case, we need $\gamma \in (0, 1)$ to ensure a well-defined return. As preliminarily outlined in~Table~\ref{tab: dtr_ai_diff2}, DTRs mainly deal with finite-horizon problems (exception made for EHRs), while JITAIs involve indefinite-horizon problems. 

\paragraph{Online and offline RL} Solving an RL task means learning an optimal way to choose the set of actions, or learning an \textit{optimal policy}, so as to maximize the expected future return. This process may follow two learning strategies: \textit{online} or \textit{offline}. In online learning, an agent learns and improves / optimizes the exploration policy $\boldsymbol{\pi}$ while following it, that is, from experiences sampled directly from $\boldsymbol{\pi}$. The policy $\boldsymbol{\pi}$ represents both the data-generating policy and the \textit{target policy}, say $\boldsymbol{d}$, the agent wants to learn about ($\boldsymbol{\pi} = \boldsymbol{d}$). However, in many decision problems, e.g., in observational settings, the agent has to learn from previously collected data. In this case, the target policy $\boldsymbol{d}$ is learned from samples collected with a policy that can be either the exploration policy (when known, e.g., in randomized studies), or, more generally, an observed or \textit{behavior policy} (that is, $\boldsymbol{\pi} \neq \boldsymbol{d}$). In the AI space, the DTR literature has predominantly focused on offline learning strategies (typically from observational data), while the mHealth domain has often adopted online RL (under a randomized setting). 

In a general RL problem, where the exploration / behavioral policy and the target policy of interest $\boldsymbol{d}$ can differ, the goal is to find an optimal policy $\boldsymbol{d}_t^* \doteq \{d^*_t\}_{\tau \geq t}$ at any time $t$, such that
\begin{align}\label{eq: opt_policy}
    \boldsymbol{d}_t^* = \argmax_{\boldsymbol{d_t}}\mathbb{E}_{\boldsymbol{d}}[\mathbf{R_t}] = \argmax_{\boldsymbol{d_t}} \mathbb{E}_{\boldsymbol{d}} \left[ \sum_{\tau \geq t} {\gamma^{\tau-t} Y_{\tau+1}} \right],
\end{align}
where the expectation is meant with respect to a trajectory distribution analogous to Eq.~\eqref{eq: P_pi}, say $P_{\boldsymbol{d}}$, with $\boldsymbol{\pi}$ replaced by a general target policy $\boldsymbol{d}$.

To estimate optimal policies, various methods have been developed so far in the RL literature (see \cite{sutton18_rl} and \cite{sugiyama2015statistical} for an overview). A traditional approach is through \textit{value functions}, which are classified into two main types: i) \textit{state-value} or simply \textit{value} functions, representing how good it is for an agent to be in a given state, and ii) \textit{action-value} functions, indicating how good it is for the agent to perform a given action in a given state.
More specifically, the time-$t$ \textit{state-value function} of policy $\boldsymbol{d}$ gives us the expected return of following policy $\boldsymbol{d}$ from time $t$ onward, conditional on history $\mathbf{h_t}$.
Formally, we denote it by $V_t^{\boldsymbol{d}}\!: \boldsymbol{\mathcal{H}_t} \to \mathbb{R}$ and define it as
\begin{align}\label{eq: value}
    V_t^{\boldsymbol{d}} (\mathbf{h_t}) \doteq
    \mathbb{E}_{\boldsymbol{d}}\left[ \mathbf{R_t} | \mathbf{H_t} = \mathbf{h_t} \right] = 
    \mathbb{E}_{\boldsymbol{d}}\left[ 
    \sum_{\tau \geq t} {\gamma^{\tau-t} Y_{\tau+1}} \middle| \mathbf{H_t} = \mathbf{h_t}
    \right],\quad \forall \mathbf{h_t} \in \boldsymbol{\mathcal{H}_t},\quad \forall t \in \mathbb{N}.
\end{align}
To ensure that the conditional expectation in $V_t^{\boldsymbol{d}} (\mathbf{h_t})$ is well defined, each history $\mathbf{h_t}\in \boldsymbol{\mathcal{H}_t}$ should have a positive probability to occur, i.e., $\mathbb{P}(\mathbf{H_t} = \mathbf{h_t}) > 0$. Note that, by definition, at time $t=0$, $V_0^{\boldsymbol{d}}(\mathbf{h_0})\doteq V_0^{\boldsymbol{d}}(x_0)$; while for the terminal time point, if any, the state-value function is $0$.

It is interesting to note that value functions define a partial ordering over policies with insightful information on the optimal ones. In fact, according to the definition of optimal policies given in Eq.~\eqref{eq: opt_policy}, a policy $\boldsymbol{d}$ is better than or equal to a policy $\boldsymbol{d}'$ if its expected return is greater than or equal (denoted as $\succeq$) to that of $\boldsymbol{d}'$ for all possible histories. Equivalently, $\boldsymbol{d} \succeq \boldsymbol{d}'$ if and only if $V_t^{\boldsymbol{d}} (\mathbf{h_t}) \geq V_t^{\boldsymbol{d}'} (\mathbf{h_t})$ for all $\mathbf{h_t} \in \boldsymbol{\mathcal{H}_t}$. As a result, optimal policies share the same (optimal) value function. Efficient estimation of the value function represents one of the most important components of almost all RL algorithms, with a central place in the decision-making paradigm. 
In DTRs, for example, evaluating the value function of a treatment regime is equivalent to evaluating the average outcome if the estimated treatment rule were to be applied to a population with the same characteristics (state or history) in the future~\citep{zhu2019proper}. Comparing the estimated value functions of different candidate treatment regimes offers a way to understand which regime may offer the greatest expected outcome.

Similar insights are given by the \textit{action-value function}. The time-$t$ \textit{action-value function} for policy $\boldsymbol{d}$, denoted by $Q_t^{\boldsymbol{d}}$, where `$Q$' stands for `Quality', is the expected return when starting from history $\mathbf{h_t}$ at time $t$, taking an action $a_t$ and following the policy $\boldsymbol{d}$ thereafter. Formally, $\forall t \in \mathbb{N}$, $Q_t^{\boldsymbol{d}}\!: \boldsymbol{\mathcal{H}_t} \times \mathcal{A}_t \to \mathbb{R}$ is defined as
\begin{align} \label{eq: q-function}
    Q_t^{\boldsymbol{d}} (\mathbf{h_t}, a_t) \doteq 
    \mathbb{E}_{\boldsymbol{d}}\left[ \mathbf{R_t} | \mathbf{H_t} = \mathbf{h_t}, A_t = a_t \right] = 
    \mathbb{E}_{\boldsymbol{d}}\left[ 
    \sum_{\tau \geq t} {\gamma^{\tau-t} Y_{\tau+1}} \middle| \mathbf{H_t} = \mathbf{h_t}, A_t = a_t \right],\quad \forall a_t \in \mathcal{A}_t,\quad \forall \mathbf{h_t} \in \boldsymbol{\mathcal{H}_t}.
\end{align}
This is also known as \textit{Q-function}, and as in Eq.~\eqref{eq: value}, $\mathbf{H_t}$ and $A_t$ are such that $\mathbb{P}(\mathbf{H_t} = \mathbf{h_t}) > 0$ and $\mathbb{P}(A_t = a_t) > 0$. 

At time $t$, the \textit{optimal value function} $V_t^* \doteq V_t^{\boldsymbol{d^*}}$ yields the largest expected return for each history with any policy $\boldsymbol{d}$, and the \textit{optimal Q-function} $Q_t^* \doteq Q_t^{\boldsymbol{d^*}}$ yields the largest expected return for each history-action pair with any policy $\boldsymbol{d}$, that is,
\begin{align} 
    Q_t^*(\mathbf{h_t}, a_t) &\doteq \max_{\boldsymbol{d_t}}Q_t^{\boldsymbol{d}}(\mathbf{h_t}, a_t), \quad \forall \mathbf{h_t} \in \boldsymbol{\mathcal{H}_t}, \forall a_t \in \mathcal{A}_t; \label{eq: optQ}\\
    V_t^*(\mathbf{h_t}) &\doteq \max_{\boldsymbol{d_t}}V_t^{\boldsymbol{d}}(\mathbf{h_t}) = \max_{a_t \in \mathcal{A}_t}Q_t^*(\mathbf{h_t}, a_t),  \ \ \forall \mathbf{h_t} \in \boldsymbol{\mathcal{H}_t}. \label{eq: optV}
\end{align}

Because an optimal action-value function is optimal for any fixed $\mathbf{h_t} \in \boldsymbol{\mathcal{H}_t}$, it follows that the optimal policy at time $t$ must also satisfy
\begin{align} \label{eq: opt_policy2}
  d^*_t(\mathbf{h_t}) 
  \in \argmax_{a_t \in \mathcal{A}_t}Q_t^*(\mathbf{h_t}, a_t).  
\end{align}

A fundamental property of the value functions used throughout RL is that they satisfy particular recursive relationships, known as \textit{Bellman} equations. For any policy $\boldsymbol{d}$, the following consistency condition, expressing the relationship between the value of a state and the values of the successor states, holds:
\begin{align} \label{eq: bellman}
    V_t^{\boldsymbol{d}} (\mathbf{h_t}) = \mathbb{E}_{\boldsymbol{d}}\left[ Y_{t+1} + \gamma V_{t+1}^{\boldsymbol{d}} (\mathbf{h_{t+1}}) \middle| \mathbf{H_t} = \mathbf{h_t} \right],\quad \forall \mathbf{h_t} \in \boldsymbol{\mathcal{H}_t},\quad \forall t \in \mathbb{N}.
\end{align}
Based on this property and Eqs.~\eqref{eq: optQ}-\eqref{eq: optV}, at each time $t$, $\forall \mathbf{h_t} \in \boldsymbol{\mathcal{H}_t}$ and $\forall a_t \in \mathcal{A}_t$, with discrete state and action spaces, the following rules, known as \textit{Bellman optimality} equations~\citep{Bellman1957}, are satisfied:
\begin{align} 
   V_t^*(\mathbf{h_t}) &= \mathbb{E} \big[ Y_{t+1} + \gamma V_{t+1}^*(\mathbf{h_{t+1}}) \mid \mathbf{H_t} = \mathbf{h_t} \big]; \label{eq: bellman_opt_V}\\
   Q_t^*(\mathbf{h_t}, a_t) &= \mathbb{E} \Big[  Y_{t+1} + \gamma \max_{a_{t+1} \in \mathcal{A}_{t+1}}Q_{t+1}^*(\mathbf{h_{t+1}}, a_{t+1}) \mid  \mathbf{H_t} = \mathbf{h_t}, A_t = a_t \Big]. \label{eq: bellman_opt_Q}
\end{align}
Here, the expectation $\mathbb{E}$ is taken with respect to the transition distribution $p_{t+1}$ only, which does not depend on the policy; thus, the subscript $\boldsymbol{d}$ can be omitted. This property allows for the estimation of (optimal) value functions recursively, from $T$ backward in time. In finite-horizon \textit{dynamic programming} (DP), this technique is known as \textit{backward induction}, and represents one of the main methods for solving the Bellman equation, also referred to as the DP equation or \textit{optimality equation}~\citep{sutton18_rl}. In infinite- and indefinite-horizon problems, using traditional backward induction is not possible, given the impossibility of extrapolating  beyond  the  time  horizon  in  the  observed data. To overcome this issue, alternative methods and additional assumptions (e.g., discounting and boundedness of rewards) are typically taken into account. Common strategies focus on time-homogeneous Markov processes to eliminate the dependence of value functions on $t$ (see e.g.,~\cite{ertefaie2018constructing,luckett2020estimating}), or revisit the Bellman optimality equation~\citep{zhou2021estimating}.

\subsection{Formalization of specific RL problems} \label{sec: MDP-MAB}
The RL problem can be posed in a variety of different ways depending on the assumptions about the level of knowledge initially available to the agent. The framework is abstract yet flexible enough to be applied to many different (sequential) problems by specifically characterizing the state and action spaces, the reward function, and other general domain (or environment) aspects, such as the time horizon or the dynamics of the process. The general framework introduced in Section \ref{Sec: full-RL} does not make any simplifying assumptions about the dependency between rewards, actions, and states: by carrying over all the available history from $t=0$, it considers a full dependency between them. We name this framework \textit{full reinforcement learning} (full-RL). 

Often, specific domains of application may have an underlying theory about the potential relationships between the key elements of an RL problem. For example, one may find it plausible to ignore the overall history and consider only the current state in the decision-making process. 
Furthermore, in some applied problems (e.g., indefinite-horizon problems), a full-RL formalization may be infeasible and/or intractable for both optimization and inference purposes. Thus, some forms of simplification in the distribution of the longitudinal histories may be needed. For example, in JITAIs, the `just-in-time' nature of decision-making requires a computationally feasible estimation and application of the decision rule continuously in time. 

Common examples of specific formalizations of an RL problem include \textit{Markov decision processes} (MDPs) and \textit{multi-armed bandit} (MAB) or contextual MAB problems. Although we discuss the MAB problem as a subclass of–or a special way of formalizing–the RL problem (as in~\cite{sutton18_rl}), we want to point out that some key researchers in the domain (see e.g.,~\cite{lattimore2020bandit}) distinguish between the two. According to them, RL is mostly associated with ML, whereas MABs are with mathematics. One driver of this choice may be related to the major focus and attention to theoretical guarantees on \textit{regret} bounds that MAB algorithms seek to satisfy. 

\begin{figure}[htb]
    \centering
    \includegraphics[width=.8\linewidth]{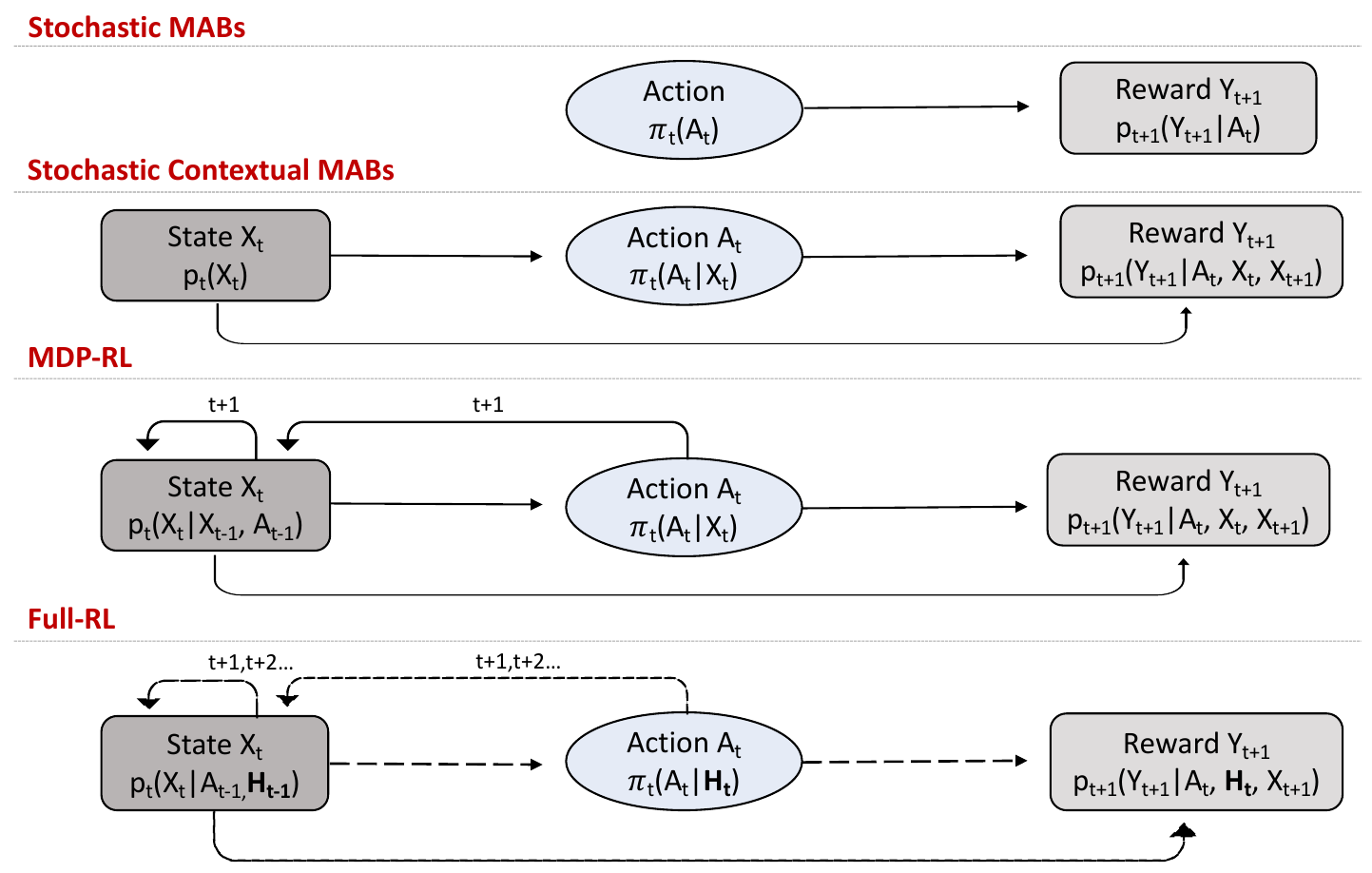}
    \caption{Graphical representation of the states, actions, and rewards relationship in a full-RL, MDP-based RL (MDP-RL), and stochastic (both contextual and context free) MAB. 
    Solid and dashed lines indicate a direct and indirect (e.g., time-delayed) effect, respectively.}
    \label{fig: mab-rl-mdp}
\end{figure}

In what follows, we illustrate these two specific formalizations, starting with the MDPs, the main framework for indefinite-horizon DTR problems. A graphical illustration of the different settings is given in Figure~\ref{fig: mab-rl-mdp}.

\subsubsection{Markov decision processes}\label{sec:MDP}
An MDP is a stochastic process used to define the dynamics of an environment and to model the interaction between the agent and the environment. It provides a convenient mathematical framework for modeling decision-making in situations where the environment is deemed to evolve according to the \textit{Markov model}~\cite{puterman2014markov}. Notably, it is the most common setting assumed in RL~\citep{van2012reinforcement}. 

What distinguishes an MDP-based RL (MDP-RL) from the full-RL framework is the environment's random memoryless characteristic.  
More specifically, assuming that the current state $X_t$ contains all the information of the past history $\mathbf{H_{t-1}}$ relevant to future predictions, it allows us to ignore the past when modeling future states and rewards. 
This property, known as \textit{Markov property}, leads to a low-dimensional representation of the past, exemplifying the trajectory distribution in Eq.~\eqref{eq: P_pi} as follows:
\begin{align*}
P_{\boldsymbol{\pi}}^{\text{MDP}} &\doteq 
p_0(x_0) \prod_{t \geq 0} \pi_{t}(a_{t} |x_t)p_{t+1}(x_{t+1}, y_{t+1}|x_t, a_{t})\\ &= 
p_0(x_0) \prod_{t \geq 0} \pi_{t}(a_{t} |x_t) p_{t+1}(x_{t+1}|x_t, a_{t})  p_{t+1}(y_{t+1}|x_t, x_{t+1}, a_{t}).
\end{align*} 
Note that under the Markov property, the agent's decisions can be entirely determined based on the current information only, as it fully determines the environment's transition-probability distributions, i.e., $p_{t+1}(\cdot, \cdot|\mathbf{H_{t}}, A_{t}) = p_{t+1}(\cdot, \cdot|X_{t}, A_{t})$, for all $t$. 
When the transition probabilities $\{p_{t+1}\}_{t \geq 0}$ are also time independent, i.e., $p_{t+1} = p$, for all $t$, the process is called \textit{time-homogeneous} or \textit{stationary} MDP. In light of this additional assumption, states, rewards, and actions are now time independent, given the information of previous time points. 
In the context of DTRs as well as JITAIs, time-homogeneous MDPs were proposed in indefinite-time horizons, as they simplify the problem by working with time-independent quantities, which do not require a backward induction strategy (see Section~\ref{DTR_infinite}).

While both full-RL and MDP-RL are typically formulated as problems with states, actions, rewards, and transition rules that depend on previous states, an exception is made for MABs, whose original formulation can be viewed as a \textit{stateless} variant of RL. 
In a typical MAB problem, either the actions and the rewards are not associated with states or they are assumed to depend only on the current state. This feature enables faster learning in settings such as JITAIs where RL is continuously implemented in an  \textit{online} fashion. This aspect will be discussed in more detail in Section~\ref{sec: divergence_DTR_JITAI}.

\subsubsection{Multi-armed bandits}\label{sec:MAB}

MAB problems, often identified as a special subclass of RL~\citep{sutton18_rl}, have a long history in statistics. They were introduced in 1933 by \cite{thompson_likelihood_1933} and extensively studied under the heading \textit{sequential design of experiments} \citep{robbins1952some, lai1985asymptotically}. 

Generally speaking, the MAB problem (also called the $K$-armed bandit problem) is a problem in which a limited set of resources (e.g., a group of individuals) must be allocated between competing choices in order to maximize the total expected reward over time. Each of the $K$ choices (i.e., \textit{arms} or actions) provide a different reward, whose probability distribution is specific to that choice. If one knew the expected reward (or value) of each action, then it would be trivial to solve the bandit problem: they would always select the action with the highest value. However, as this information is only partially gained for the selected actions, at each decision time $t$ the agent must trade-off between optimizing its decisions based on acquired knowledge up to time $t$ (\textit{exploitation}) and acquiring new knowledge about the expected rewards of the other actions (\textit{exploration}). 

MAB strategies were originally proposed to solve stateless problems, in which the reward depends uniquely on actions. Subsequently, a `stateful' variant of MABs, named \textit{contextual} MAB (C-MAB), in which actions are associated with some state, or \textit{context}, was introduced. However, unlike full-RL and MDP-RL, in contextual MABs, actions do not have any effect on the next states. In addition, generally, there are no transition rules from one state to another in subsequent times. This implies that states, actions, and rewards can be treated as a set of separate events over time.
The most typical assumption is that contexts $\{X_t\}_{t \in \mathbb{N}}$ are independent and identically distributed (IID) with some fixed but unknown distribution. This means that action $A_t$ at time $t$ has an \textit{in-the-moment} effect on the proximal reward $Y_{t+1}$ at time $t+1$, but not on the distribution of future rewards $\{Y_\tau\}_{\tau \geq t+2}$, for which the IID property holds as well. Under this assumption, one can be completely myopic and ignore the effect of an action on the distant future in searching for a good policy. This problem is better known as \textit{stochastic} MABs, in contrast to \textit{adversarial} MABs~\cite{lattimore2020bandit}, in which no independence assumptions are made on the sequence of rewards.
In stochastic contextual MABs, and further in the context-free MAB problem, the trajectory distributions are simplified as follows:
\begin{align} \label{eq:traj_MAB} 
P_{\boldsymbol{\pi}}^{\text{C-MAB}} &\doteq 
p_0(x_0) \prod_{t \geq 0} \pi_{t}(a_{t} |x_t)p_{t+1}(x_{t+1}, y_{t+1}|x_t, a_{t}) = p_0(x_0) \prod_{t \geq 0} \pi_{t}(a_{t} |x_t)
p_{t+1}(x_{t+1}) p_{t+1}(y_{t+1}|x_t, x_{t+1}, a_{t}); \\
P_{\boldsymbol{\pi}}^{\text{MAB}} &\doteq 
\prod_{t \geq 0} \pi_{t}(a_{t})p_{t+1}(y_{t+1}|a_{t}). \nonumber
\end{align}

Note that, since the effect of an action in the stochastic MAB is in-the-moment, the bandit problem is formally equivalent to a one-step/state MDP, wherein the states progression is not taken into account. Thus, compared to MDP-RL and full-RL, MABs provide a simplified structure of the relationships between the components of RL within time. For a graphical summary, see Figure~\ref{fig: mab-rl-mdp}. 

As in the general RL problem, the goal of an MAB problem is to select the optimal arm at each time $t$ so as to maximize the expected return, alternatively (and with a slightly different nuance) expressed in the bandit literature in terms of minimizing the \textit{total regret}. Indeed, in (online) real-world problems, until we can identify the best (unique) arm, we need to make repeated trials by pulling the different arms. The loss that we incur during this learning phase (i.e., the time spent for learning the best arm) represents what is called regret, i.e., how much we regret not picking the best arm. Formally, denoted by $A_t^* \doteq \argmax_{a_t \in \mathcal{A}}\mathbb{E}(Y_{t+1} | X_t = x_t, A_t = a_t)$
the optimal arm at time $t$, we define the \textit{immediate regret} $\Delta(A_t)$ of action $A_t$ as the difference between the expected reward of the optimal arm $A_t^*$ and the expected reward of the ultimately chosen arm $A_t$, i.e.,
\begin{align} \label{eq: regret}
    \Delta(A_t) \doteq \mathbb{E}(Y_{t+1} | X_t, A_t^*) - \mathbb{E}(Y_{t+1} | X_t, A_t).
\end{align}
Given a horizon $T$, the goal of the learner is to minimize the total regret given by $\text{Reg}(T) \doteq \sum_{t = 0}^T \Delta(A_t)$. Note that the agent may not know ahead of time how many time points $T$ are to be played. Therefore, the goal is to perform well not only at the final time point $T$, but also during the learning phase. For example, in a dose-finding problem as the one mentioned in Section \ref{sec:MDP}, the objective may not only be to minimize the sum of toxicities over time, but also to ensure that these toxicities have a proper upper limit–thus, limiting extremely harmful adverse events–uniformly over time. For this reason, as we will see later in Section \ref{Sec: JITAIs_mHealth}, theoretical works on regret bounds occupy a central place in the bandit literature.

\subsection{RL and AIs: a joint overview} \label{sec: RL-AIs-joint}

So far, we have introduced the RL as a mathematical framework for sequential decision-making problems and discussed its characterization in illustrative AI examples of interest. Before diving deep into the rich literature of existing RL methods for building (optimized) AIs, we provide the reader with a joint overview of the different problems, which notably share the same key elements and a common optimization objective.
\begin{table*}[b]
\centering
\small
    \caption{Notation and terminology of reference of the key elements in RL, MAB, DTR and JITAI problems}
    \begin{tabu} to \textwidth { X[0.5] | X[0.85] X[0.85] X[1.4] X[1.1]}
        \multicolumn{1}{c}{\textbf{Notation}} & \multicolumn{4}{c}{\textbf{Terminology}}\\
        \hline
        & RL & MABs & DTRs & JITAIs \\
        \hline
        $i$& Trajectory & Trajectory & Patient & User\\
        $t$& Time point & Round, Time point & Stage, Interval, Time point & Time point\\
        $X$& State & Context & Tailoring variable & Contextual variable\\
        $A$& Action & Arm & Treatment, Intervention & Intervention option\\
        $Y$& Reward & Reward & Intermediate outcome & Proximal outcome\\
        $\mathbf{H}$ & History & History, Filtration & History & History\\
        $\boldsymbol{\pi}$, $\boldsymbol{d}$& Policy & Policy  & (Dynamic) treatment regime & Policy\\
        $\boldsymbol{\pi}^*$, $\boldsymbol{d}^*$& Optimal policy & Optimal policy & Optimal DTR & Optimal policy\\
        \hline
    \end{tabu}
    \label{tab: terminology}
\end{table*}
As such, they can be unified under a unique formal framework and solved with techniques developed under the RL paradigm.

Table~\ref{tab: terminology} outlines the terminologies of reference in each setting, with a unified notation adopted from the general RL. Note that, while we report only the most common terminology employed in each setting, lexical borrowing is widely used across the different theoretical and applied domains. To illustrate, the term `treatment policy', or just `policy' is often used in place of `treatment regime' in the DTR literature. Also note that, in general, the terminology adopted in a specific application is guided by the RL method and framework used in that application; see e.g., the similarity between the terms used in JITAIs and MABs such as `contextual variables' and `context' (i.e., the state of the environment). Both contextual and tailoring variables represent the set of baseline and time-varying information that is used to personalize decision-making. Alternative terms such as covariates or features (which we use with slightly different meaning, as we discuss in Section \ref{Sec: LinUCB}) are also common. To help the reader navigate the different terminologies, an extended version of Table~\ref{tab: terminology}, detailing all the notation and acronyms used in the main manuscript, is provided in Supplementary Material B.

We anticipate that most (if not all) of the methods to construct JITAIs would generally belong to the MAB class, although the applied literature commonly refers to it with the generic `reinforcement learning' name (see e.g.,~\cite{yom2017encouraging,liao2020personalized,figueroa2020guidelines}). In DTRs, the predominant class of methods is full-RL, followed by MDP-RL proposed specifically for indefinite-horizon (e.g., EHR-based) DTR problems. In fact, the underlying theory of DTRs--characterized by potential delayed or carried-over effects of treatment over time--and the importance of the evolving history of a patient for predicting future outcomes requires accurate consideration of information from previous time points. Generally, the meaningful relationship between the different variables of a patient's history does not allow simplifying or ignoring the (state-)transition rules, making full-RL (and occasionally MDP-RL) the ideal option. On the other hand, the behavioral theory of a momentary effect of an intervention on the proximal outcome makes MABs a suitable framework in mHealth settings. In addition, the reduced computational burden from carrying through all the historical information allows MAB strategies to be applied continuously in time, e.g., every hour, and efficiently construct JITAIs.

\section{A Survey of RL Methods for Adaptive Interventions} \label{Sec: RLmethods}

Methodology for constructing \textit{optimal} AIs, i.e., the ones that, if followed, would yield the most favorable (typically long-term) mean outcome, is of considerable interest within the domain of precision medicine, and comprises a large body of research within theoretical and applied sciences \citep{chakraborty_statistical_2013, laber_dynamic_2014,kosorok_precision_2019}. Although their relevance has been long documented within statistics and causal inference (see Section~\ref{Sec: DTRs_excursus}), recently it has generated a lot of interest within the computer science and engineering communities, due to the similarity between the mathematical formalization of AIs and the RL framework.

\subsection{RL methods for dynamic treatment regimes} \label{Sec: DTRs}

\subsubsection{A historical overview} \label{Sec: DTRs_excursus}

Perhaps due to the need to identify causal relationships, the study of AIs originated in causal inference with the pioneering works of Robins (see e.g.,~\cite{robins1986new, robins1994correcting} for DTRs). Over an extended period of time, the author introduced three basic approaches for finding effects of time-varying regimes in the presence of confounding variables: the parametric \textit{G-formula} or \textit{G-computation}~\citep{robins1986new}, \textit{structural nested mean models} with the associated method of \textit{G-estimation}~\citep{robins1989analysis, robins1992estimation, robins1994correcting}, and \textit{marginal structural models} with the associated method of \textit{inverse probability of treatment weighting} (IPW)~\citep{robins2000marginal}. 

A number of methods have subsequently been proposed within statistics, including both frequentist and Bayesian approaches~\citep{thall2000evaluating, thall2002selecting, thall2007bayesian, lavori_design_2000}. However, all estimate the optimal DTR based on distributional assumptions on the data-generation process via parametric models, and, as such, can easily suffer from model misspecification~\citep{zhao2015new}. The first semiparametric method for estimating optimal DTRs was proposed by Murphy~\cite{murphy_optimal_2003}, immediately followed by Robins~\cite{robins2004optimal}, who introduced two alternative approaches using G-estimation. 
These methods use \textit{approximate dynamic programming}, where `approximate' refers to the use of an approximation of the value or Q-function introduced in Eq.~\eqref{eq: q-function}, or parts thereof. Thus, they can be considered as the first prototypes of RL-based approaches in the AIs literature. 

RL methods represent an alternative approach to estimating DTRs that have gained popularity due to their success in addressing challenging sequential decision-making problems, without the need to fully model the underlying generative distribution. The connection between statistics and RL (previously confined to the computer science and control theory literature) was bridged by Murphy~\cite{murphy2005generalization}, who proposed estimating optimal DTRs with Q-learning~\citep{watkins89_rl, sutton18_rl}. 
Promptly, a large body of research has embraced the use of Q-learning, integrating various parametric, semiparametric, and nonparametric strategies \citep{murphy2005generalization, chakraborty_statistical_2013, chakraborty_dynamic_2014, laber2014interactive} to model the Q-function. 
Q-learning and the semiparametric strategies of Murphy~\cite{murphy_optimal_2003} and Robins~\cite{robins2004optimal} are considered \textit{indirect methods}: optimal DTRs are indirectly obtained by first estimating an optimal objective function (e.g., the Q-function), and then getting the associated (optimal) policy. In contrast, IPW-based strategies~\citep{robins2000marginal, murphy2001marginal, wang2012evaluation} seek optimal policies by directly looking for the policy (within a prespecified class of policies) that maximizes an objective function (e.g., the expected return), without postulating an outcome model \citep{zhao2012estimating}; they are regarded as \textit{direct methods}. 

In what follows, we review existing RL techniques for developing DTRs focusing on the indirect methods, while an up-to-date review including direct methods can be found in~\citep{deliu_dynamic_2022}. We cover both finite- and indefinite-horizon settings.  
We emphasize that most of the current work in DTRs deals with finite-horizon problems and \textit{offline learning} procedures that assume access to a collection of observed trajectories. This opposes to the JITAIs tradition–originated with the practical need to deliver AIs in real time–which uses an \textit{online learning} approach for performing data collection and policy optimization simultaneously. Such procedures can be deployed indefinitely, conditional on practical limitations.
In DTRs, the indefinite-horizon setting, particularly suitable for chronic diseases where the number of stages 
can be arbitrarily large, has been addressed only recently. Nevertheless, it remains relatively understudied. 

\subsubsection{Finite-horizon DTR problems} \label{Sec: DTR_finite}
Finite-horizon DTR problems are designed to identify optimal treatment policies $\boldsymbol{d}^* = \{d_t^*\}_{t = 0,\dots,T}$ over a fixed and known period of time $T < \infty$. Learning methods typically use \textit{offline} RL based on finite (experimental or observational) data trajectories of a sample of say $N$ patients, and causal assumptions about the data (see e.g.,~\cite{deliu_dynamic_2022}). Each patient trajectory has the form $(X_0, A_0, Y_1, \dots, X_T, A_T, Y_{T+1})$, with $X_0$ and $X_1,\dots,X_T$ the pretreatment and evolving information, respectively, $A_0,\dots,A_T$ the assigned treatments, and $Y_1,\dots,Y_{T+1}$ the intermediate outcomes. When a single distal (end-of-study) outcome $\overset{\infty}{Y}$ is considered, all intermediate quantities $\{Y_t\}_{t=1,\dots,T}$ are taken as $0$, and $Y_{T+1} = \overset{\infty}{Y}$, as discussed in Section~\ref{Sec: RLforAIs}. 
Note that, especially in observational settings, the decision points $t$ can exhibit greater variability and display less consistent patterns across subjects. In this case, the specific time points $t$ can substantially differ among different individuals, requiring an accurate definition of the admissible time intervals, therefore the $N$ data trajectories, when conducting a DTR analysis. 

In finite-horizon problems, RL methods are mainly based on DP or approximate DP procedures. These include Q-learning~\citep{murphy2005generalization}, with the Q-function as the objective, and A-learning~\citep[]{robins2004optimal, murphy_optimal_2003}, which focuses on contrasts of conditional mean outcomes. 
We now discuss the former, assuming throughout this section deterministic policies, that is, policies that map histories $\mathbf{h}$ directly into actions or decisions, i.e., $\boldsymbol{d}(\mathbf{h}) = \mathbf{a}$.

\paragraph*{Q-learning with function approximation.}

In Section \ref{Sec: RL_framework}, we showed that optimal value functions can be obtained by iteratively solving the Bellman optimality relationship in Eqs.~\eqref{eq: bellman_opt_V}-\eqref{eq: bellman_opt_Q}. In finite-horizon DP problems, this procedure is known as backward induction. However, the iterative process may be memory and computationally intensive, especially for large state and action spaces. Furthermore, traditional DP procedures assume an underlying model for the environment, which is often unknown due to unknown transition probability distributions. 
Q-learning~\citep{watkins89_rl} offers a powerful and scalable tool to overcome the modeling requirements as well as the computational burden of traditional DP-based methods and constitutes the core of modern RL.

The general idea of Q-learning is that, at each new $t$, the Q-function is updated based on a previous value and the new acquired information:
\begin{align*}
    Q_t^{\boldsymbol{d}}(\mathbf{h_t}, a_t) \leftarrow Q_t^{\boldsymbol{d}}(\mathbf{h_t}, a_t) + \alpha_t \Big[Y_{t+1} +\gamma \max_{a_{t+1} \in \mathcal{A}_{t+1}} Q_{t+1}^{\boldsymbol{d}}(\mathbf{h_{t+1}}, a_{t+1})- Q_t^{\boldsymbol{d}}(\mathbf{h_t}, a_t) \Big],
\end{align*}
with $\alpha_t$ a constant that determines to what extent the newly acquired information overrides the old information or how fast learning takes place, and $\gamma$ a discount factor that balances immediate and future rewards (in finite-horizon problems, it is generally set to one). 

The original version of this approach is known as \textit{tabular Q-learning}~\cite{sutton18_rl}. This is based on storing the Q-function values for each possible state and action in a lookup table and choosing the one with the highest value. Since the agent selects the actions based on their maximum associated Q-function value, this is equivalent to \textit{exploiting} (recall the notion of exploitation introduced in Section~\ref{sec:MAB}). However, the tabular approach is slow and impractical for large state and action spaces. A powerful and scalable solution to this problem is a more recent version of Q-learning, known as \textit{Q-learning with function approximation}~\citep{sutton18_rl, murphy2005generalization}. This version first assumes an approximation space for each of the Q-functions in Eq.~\eqref{eq: q-function}, e.g., $\mathcal{Q}_t \doteq \left \{Q_t^{\boldsymbol{d}}(\mathbf{h_t}, a_t; \theta_t)\!: \theta_t \in \Theta_t \right \}$, with parameter space $\Theta_t$ a subset of the Euclidean space, and then estimates the optimal stage-$t$ Q-functions $Q^*_t$ backward in time for $t = T, T-1, \dots, 0$~\citep{bather2000decision}. According to~Eq.~\eqref{eq: opt_policy2}, estimating an optimal regime $\hat{\boldsymbol{d}}^* = (d_0^*(x_0), d_1^*(\mathbf{h_1}), \dots, d_T^*(\mathbf{h_T}))$ is equivalent to getting estimates of the optimal Q-functions, or in this case, getting an estimate $\hat{\theta}_t$, $t = 1,\dots,T$, of the parameters, i.e.,
\begin{align*}
    \hat{d}_t^*(\mathbf{h_t}) = \argmax_{a_t \in \mathcal{A}_t} \hat{Q}_t^*(\mathbf{h_t}, a_t) \doteq \argmax_{a_t \in \mathcal{A}_t} Q_t^*(\mathbf{h_t}, a_t; \hat{\theta}_t) \doteq d_t^*(\mathbf{h_t}; \hat{\theta}_t).
\end{align*}
Noticing, for example, that the Q-function is a conditional expectation, we can get the optimal Q-functions as:
\begin{align*} 
    Q_{t}^*(\mathbf{h_{t}}, a_{t}; \hat{\theta}_{t}) \doteq \hat{\mathbb{E}}_N[Y_{t} + \max_{a_{t+1} \in \mathcal{A}_{t+1}} Q_{t+1}^*(\mathbf{h_{t+1}}, a_{t+1}; \hat{\theta}_{t+1}) \mid \mathbf{H_{t}} = \mathbf{h_{t}}, A_{t} = a_{t}],
\end{align*}
with $\hat{\mathbb{E}}_N$ denoting the empirical mean over a sample of $N$ units. The procedure is illustrated in Supplementary Material E, and a more specific implementation with linear regression is given in Supplementary Material C.

It is important to recognize that the estimated regime $\mathbf{\hat{d}}^*$ may not be a consistent estimator for the true optimal regime $\boldsymbol{d}^*$, unless all models for the Q-functions are correctly specified. 
A strategy that may offer robustness to Q-function misspecification is A-learning~\citep{robins2004optimal, murphy_optimal_2003}, where `A' stands for the `advantage' incurred if the optimal treatment were given as opposed to what was actually given. A-learning represents a class of alternative methods to Q-learning, predicated on the fact that it is not necessary to specify the entire Q-function to estimate an optimal regime. A more in-depth discussion is provided in Supplementary Material D. Schulte et al.~\cite{schulte2014q} showed that A-learning outperforms Q-learning under misspecifications of Q-function models.  

Given that a linear regression model may be quite simple and prone to misspecification, more sophisticated approximators can be used both in Q-learning and in A-learning. These include \textit{support vector regression}~\citep{zhao_reinforcement_2009} and \textit{deep neural networks}~\citep{atan2018deep}, among others. 

\begin{figure}[htb]
    \centering
    \includegraphics[width=.8\linewidth]{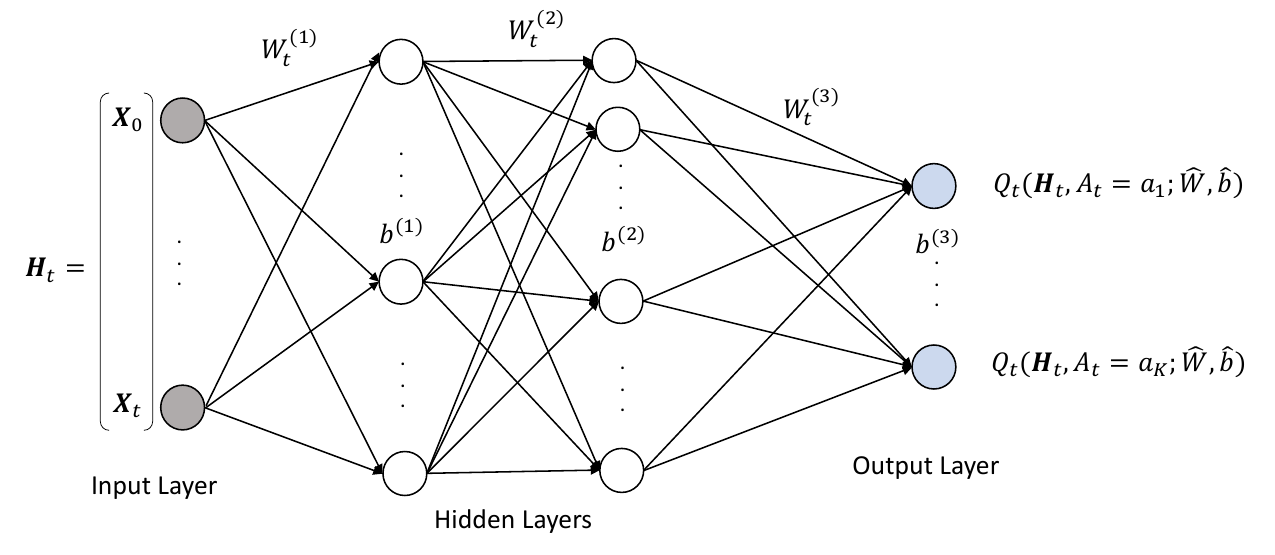}
    \caption{Schematic of a feed-forward neural network. It is characterized by a set of neurons, structured in four layers ($L=4$), where each neuron processes the information forward from one layer to the next one. Information is nonlinearly transformed according to unknown weights $W^{(l)}$ and bias $b^{(l)}$ parameters, $l=1,\dots,L-1$.}
    \label{fig: ff-nn}
\end{figure}

\paragraph*{Deep Q-network.}
The tremendous success achieved in recent years by RL has been greatly enabled by the use of advanced function approximation techniques such as deep neural networks (DNNs)~\citep{jonsson2019deep, silver2017mastering, mnih2015human}, giving rise to the \textit{deep Q-network} algorithm~\citep{mnih2015human}. Specifically, at a given time $t$, a DNN (see~\cite{goodfellow2016deep} for an overview of existing DNN architectures) is used to fit a model for the Q-function in a supervised way and then estimate the optimal Q-function: histories $\{\mathbf{H}_{t,i}\}_{i=1,\dots,N}$ are given as input, and the predicted Q-function values $Q_t^{\boldsymbol{d}}(\mathbf{H}_{t}, a_{t}; \hat{\mathbf{W}},\hat{\mathbf{b}})$ associated with each action $a_{t} \in \mathcal{A}_t$, e.g., with $\mathcal{A}_t = \{a_1,\dots,a_K\}$, are generated as output. $\mathbf{W}$ and $\mathbf{b}$ represent the unknown \textit{weight} and \textit{bias} parameters of a typical DNN; see e.g., the schematic of a \textit{feed-forward neural network} in Figure~\ref{fig: ff-nn}.

Once Q-function estimates are obtained with the DNN, the algorithm proceeds with executing, in an emulator, an action according to an exploration scheme named $\epsilon$-greedy~\citep{sutton18_rl}. This probabilistically chooses between the optimal action so far (i.e., the one with the highest estimated Q-function value) and a random action. Specifically, $\epsilon$ is the \textit{exploration} probability for a random action. At the end of the execution sequence, first the Q-function is re-estimated based on the observed reward, and then the DNN parameters are updated using the last Q-function estimates. The pseudo-algorithm is given in Supplementary Material E.

A DNN offers a more flexible and scalable approach, particularly suitable for real-life complexity, high dimensionality, and high heterogeneity. Compared to their shallow counterparts, they enable automatic feature representation and can capture complicated relationships (see e.g., the application in the graft-versus-host disease of \cite{liu2017deep}). A general limitation of indirect methods such as Q-learning, is that the optimal DTRs are estimated in a two-step procedure: first, the Q-functions are estimated using the data, and then these are optimized to infer the optimal DTR. In the presence of high-dimensional information, even with flexible nonparametric techniques such as DNNs, it is possible that these conditional functions are poorly fitted, with the derived DTR far from optimal. Furthermore, as demonstrated by \cite{zhao2012estimating}, indirect approaches may not necessarily result in the maximum long-term clinical benefit, motivating direct methods. We refer to~\cite{deliu_dynamic_2022,tsiatis_dynamic_2021} for a survey of direct approaches. 

Nonetheless, we emphasize that here we present indirect methods in some detail because they are somewhat similar to well-known regression methods that most readers can relate to. Furthermore, many of the methods for developing JITAIs (e.g., Thompson sampling, among the other methods discussed in Section~\ref{Sec: JITAIs_mHealth}) are also regression-based methods. Thus, by focusing on regression-type methods across apparently disjoint application domains, we help enhance the synergy between them.

\subsubsection{Indefinite-horizon DTR problems}\label{DTR_infinite}
While in computer science there is a vast literature on estimating optimal policies over an increasing time horizon \citep{szepesvari2010algorithms, sugiyama2015statistical}, that is not the case in DTRs. In fact, by adopting backward induction, most existing methods cannot extrapolate beyond the time horizon in the observed data. Nevertheless, for some chronic conditions or those with very short time steps, including mHealth applications (see Section \ref{Sec: RLforAIs}), the time horizon is not definite. Treatment decisions are made continuously throughout the life of a patient, with no fixed time point for the final treatment decision. 

To the best of our knowledge, only a limited number of statistical methodologies have been developed for the indefinite-horizon setting. These include the indirect \textit{greedy gradient Q-learning} method of \cite{ertefaie2018constructing}, and the direct \textit{V-learning} approach of \cite{luckett2020estimating}, who proposed to search for an optimal policy over a prespecified class of policies. 
More recently, a minimax framework called proximal temporal consistency learning was proposed~\citep{zhou2021estimating}. We now detail the first two approaches, while for the third, we refer the reader to the original work in~\cite{zhou2021estimating}.

\paragraph*{Greedy gradient Q-learning.} The first extension to indefinite-time horizons in DTRs was proposed in \cite{ertefaie2018constructing}, under the time-homogeneous Markov assumption (see Section \ref{sec:MDP}). Although not imposed by general DTR methods, such assumption overcomes the need for backward induction, and exemplifies inference by working with time-independent Q-functions. 

We adopt the notation of the previous sections and introduce an absorbing state $c$, representing a loss-to-follow-up, e.g., death, event. We assume that at each time $t$, covariates $X_t$ take values in a finite state space $\mathcal{X}^\star \doteq \mathcal{X} \cup \{c\}$, with $\mathcal{X} \cap \{c\} = \emptyset$. Let the action space $\mathcal{A}_{x}$ be finite and defined by covariate information such that $\mathcal{A}_{x}$ consists of $ 0 < K_{x} \leq K$ treatments, with $K$ being the total number of treatments over the time horizon. For any $t$ such that $X_t = c$, let $A_{x} = \mathcal{A}_c = \{u\}$, where $u$ stands for `undefined'. Now, denoting a stopping time (e.g., death) by $\widetilde{T} \doteq \inf \{t > 0: X_t = c\}$, individual trajectories are of the form $(X_0, A_0, R_1,\dots, X_{\widetilde{T}-1}, A_{\widetilde{T}-1}, R_{\widetilde{T}}, X_{\widetilde{T}})$. Note that $\mathbb{P}(\widetilde{T} < \infty | X_0, A_0) = 1$, regardless of $(X_0, A_0)$. 
Based on these specifications, the indefinite time-$t$ Q-function for regime $\boldsymbol{d}(\mathbf{h_t}) = \boldsymbol{d}(x_t) = \boldsymbol{d}(x)$, for $x \in \mathcal{X}$, is given by:
\begin{align*}
    Q^{\boldsymbol{d}} (x, a) \doteq 
    \mathbb{E}_{\boldsymbol{d}}\left[ \mathbf{R_t} | X_t = x_t, A_t = a_t \right] = 
    \mathbb{E}_{\boldsymbol{d}}\left[ 
    \sum_{\tau = 0}^{\infty}{\gamma^{\tau-t} Y_{\tau+1}} \middle| X_t = x_t, A_t = a_t \right].
\end{align*}
We set $Q^*(c,a) = 0$ since the return is $0$ after an individual is lost to follow-up.

For estimating an optimal DTR, Q-learning is proposed. Let $Q(x,a; \theta^*)$ be a parametric model for $Q^*(x, a)$ indexed by $\theta^* \in \Theta \subseteq \mathbb{R}^n$, with $n\geq 1$, and postulate a linear model with interactions, i.e., $Q(x,a; \theta^*) = \theta^{*T}f(x,a)$, with $f(x,a)$ being a known feature vector summarizing the state and treatment pair. To ensure $Q^*(c,a) = 0$, we also need $f(c,a) = 0$. Now, defining $f(X_t,A_t) \doteq \nabla_{\theta^*} Q (X_t, A_t; \theta^*)$, with $\nabla$ the gradient, Bellman optimality suggests and motivates the following unbiased estimating function for $\theta^*$:
\begin{align} \label{GGQ}
    \hat{D}(\theta^*) = \hat{\mathbb{P}}_N \Bigg\{ 
    \sum_{t=0}^{T-1} \Bigg(Y_{t+1} + \gamma \max_{a \in \mathcal{A}_{X_{t+1}}}Q(X_{t+1},a; \theta^*) - Q(X_t, A_t; \theta^*) \Bigg) f(X_t,A_t)
    \Bigg \}.
\end{align}

Note that the estimating function in Eq.~\eqref{GGQ} is a nonconvex and nondifferentiable function of $\theta^*$, which complicates the estimation process. Under regularity conditions, the authors suggested that any solution $\hat{\theta}^*$ can be equivalently defined as a minimizer of $\hat{M}(\theta^*) \doteq \hat{D}(\theta^*)^T\hat{S}^{-1}\hat{D}(\theta^*)$, with $\hat{S} \doteq \hat{\mathbb{P}}_N \left\{\sum_{t=0}^{T-1}f(X_t, A_t)^{\otimes 2} \right\}$, and $x^{\otimes 2} \doteq xx^T$, for any vector $x$. If $\hat{\theta}^* = \argmin_{\theta^* \in \Theta}\hat{M}(\theta^*) $ is the unique solution, then $\hat{Q}^*(x,a) = Q(x,a; \hat{\theta}^*)$, and the corresponding optimal regime is given by $\hat{\boldsymbol{d}}^* = \argmax_{a \in \mathcal{A}_x}Q(x,a; \hat{\theta}^*)$.  

\paragraph*{V-learning.} The greedy gradient Q-learning approach based on Eq.~\eqref{GGQ} involves a nonsmooth max operator that makes estimation difficult without large amounts of data~\citep{laber2014interactive, linn2017interactive}. Motivated by an mHealth application, where policy estimation is continuously updated in real time as data accumulate (starting with small sample sizes), an alternative method is proposed in~\cite{luckett2020estimating}. Under the same time-homogeneous MDP assumption, provided that interchange of the sum and integration is justified, the authors consider the value function
\begin{align*}
    V_t^{\boldsymbol{d}}(x_t) = \sum_{\tau \geq t} \mathbb{E} \left[ \gamma^{\tau-t}Y_{\tau+1} \left( \prod_{v = t}^\tau \frac{d(A_v|X_v)}{\pi_v(A_v| X_v)}\right) \middle | X_t = x_t \right],
\end{align*}
and follow a direct approach to directly maximize estimated values over a prespecified class of policies. In light of the Bellman equation in Eq.~\eqref{eq: bellman}, it follows that, for any function $f$ defined on the state space $\mathcal{X}_t$, the following importance-weighted variant is satisfied: 
\begin{align*}
    0 = \mathbb{E} \left[\frac{d(A_t|X_t)}{\pi_t(A_t| X_t)} \left(Y_{t+1} + \gamma V^{\boldsymbol{d}}(X_{t+1}) - V^{\boldsymbol{d}}(X_t)\right)f(X_t) \right].
\end{align*}

Let $V^{\boldsymbol{d}}(x;\theta)$, with $ \theta \in \Theta \subseteq \mathbb{R}^n $, be a model for $V^{\boldsymbol{d}}(x)$. Assume that $V^{\boldsymbol{d}}(x;\theta)$ is differentiable everywhere in $\theta$ for fixed $x$ and $\boldsymbol{d}$. Then, the proposed estimating equation is given by:
\begin{align*}
    \hat{\Lambda}(\theta) = \hat{\mathbb{P}}_N \Bigg[\sum_{t=0}^T \frac{d(A_t|X_t)}{\pi_t(A_t| X_t)} \Bigg( Y_{t+1} + \gamma V^{\boldsymbol{d}}(X_{t+1}; \theta) - V^{\boldsymbol{d}}(X_t; \theta) \Bigg) \nabla_{\theta} V^{\boldsymbol{d}}(X_t; \theta)\Bigg].
\end{align*}

Again, $\hat{\theta}$ can be obtained by minimizing $\hat{M}(\theta) \doteq \hat{\Lambda}(\theta)^T\hat{S}^{-1}\hat{\Lambda}(\theta) + \lambda \mathcal{P}(\theta)$, with $\hat{S}$ a positive definite matrix in $\mathbb{R}^{n\times n}$, $\lambda$ a tuning parameter, and $\mathcal{P}: \mathbb{R}^n \to \mathbb{R}_+$ a penalty function. 
The estimated optimal regime $\hat{d}^*$ is the argmax of $V^{\boldsymbol{d}}(x;\hat{\theta})$. Compared to greedy gradient Q-learning, V-learning requires modeling both the policy and the value function, but not the data-generating process. In addition, by directly maximizing the estimated value over a class of policies (see~\cite{luckett2020estimating} for more details), it overcomes the issues of the nonsmooth max operator in Eq.~\eqref{GGQ}. The method is applicable over indefinite horizons and is suitable for both offline and online learning, which is typical in JITAIs.

\subsection{JITAIs in mHealth} \label{Sec: JITAIs_mHealth}

Unlike DTRs, where the number of decision points is generally small, JITAIs are defined upon a random and indefinitely large number of times. They are carried out in dynamic environments with the scope of capturing rapid changes in an individual user's context and needs~\citep{nahum2015building, nahum2018just}. 
Methodologies for optimizing JITAIs require the ability to learn nearly continuously, with no definite time horizon. Furthermore, learning is performed \textit{online} as data accumulate, often using trajectories defined over very short time periods. Note that in such settings, the exploration policy $\boldsymbol{\pi}$ used to collect the samples corresponds to the target policy $\boldsymbol{d}$ we want to improve and optimize; that is, $\boldsymbol{\pi} = \boldsymbol{d}$. Thus, existing methods for DTRs, which mainly target a finite-time horizon problem and are implemented \textit{offline} (e.g., Q-learning), are not directly applicable to JITAIs. Furthermore, by carrying over an entire history of an individual, they may not be feasible from a computational perspective. 

As discussed in Section~\ref{Sec: RL_framework}, the standard approach for developing JITAIs is given by contextual MABs~\citep{tewari2017ads}, an intermediate solution between MABs~\citep{auer2002nonstochastic} and the full-RL approach used in DTRs. With a few exceptions, contextual MAB algorithms applied in mHealth rely on two fundamental bandit strategies, originally implemented in advertising: the \textit{linear upper confidence bound} (LinUCB)~\citep{li2010contextual} and the \textit{linear Thompson sampling} (LinTS)~\citep{agrawal2013thompson}.

\subsubsection{Contextual MABs with LinUCB exploration} \label{Sec: LinUCB}

So far, optimal AIs have typically been identified by finding the optimal Q-functions recursively with the Bellman relationships. In contrast, LinUCB~\citep{li2010contextual, chu2011contextual} 
employs the underlying idea of MABs, where the optimal policy is the set of the optimal stage-$t$ arms, for all $t$, defined individually for each time $t$ as: $d_t^* \doteq A_t^* \doteq \argmax_{a_t \in \mathcal{A}}\mathbb{E}(Y_{t+1} | X_t = x_t, A_t = a_t)$. Notice that this objective function represents a myopic version of the Q-function in Eq.~\eqref{eq: q-function} by taking $\gamma=0$, and reflects a stochastic MAB setting where contexts are IID and one can ignore the previous history given the last state $x_t$ (see also Eq.~\eqref{eq:traj_MAB}); that is,
\begin{align}
    Q_t^{\boldsymbol{\pi}} (\mathbf{h_t}, a_t) = Q_t^{\boldsymbol{\pi}} (x_t, a_t) =  
    \mathbb{E}\left[ Y_{t+1} \middle| X_t = x_t, A_t = a_t \right],\quad \forall a_t \in \mathcal{A}_t, \quad \forall x_t \in \mathcal{X}_t, \quad \forall t \in \mathbb{N}. \nonumber
\end{align}

The specific solution of LinUCB is based on performing an efficient exploration by favoring arms for which a confident value has not been estimated yet and avoiding arms which have shown a low reward with high confidence. This confidence is measured by the \textit{upper confidence bound} of the expected reward value for each arm. The underlying assumption is that the conditional expected reward is a linear function of a context-action feature $f$, that is, $\mathbb{E}[Y_{t+1} | X_t, A_t] = f(X_t, A_t)^T\mu$, with $\mu \in \mathbb{R}^{n}$ the unknown coefficient vector associated with the feature $f$. In this work, we consider general features $f$ (constructed via linear basis, polynomials or splines expansion, among others; see e.g.,~\cite{marsh2001spline}) rather than a standard linear function that may fail to capture nonlinearities in the data.

Under the linear model assumption, the LinUCB idea is to estimate at each time $t$ an upper bound, say $U_t(a_t)$, for the expected reward of each arm $a_t$. The LinUCB estimator is defined as
\begin{align} \label{eq: UCB}
    \hat{U}_t(a_t) \doteq \hat{Q}_t^{\boldsymbol{\pi}}(x_t, a_t) + \alpha s_t(a_t)  = f(X_t = x_t, A_t = a_t)^T \hat{\mu}_{t} + \alpha s_t(a_t),
\end{align}
where $\alpha > 0$ is a tuning parameter that controls the trade-off between exploration and exploitation: small values of $\alpha$ favor exploitation while larger values of $\alpha$ favor exploration. The first part $f(X_t, A_t = a_t)^T \hat{\mu}_{t}$, with $\hat{\mu}_{t} \doteq B_t^{-1}b_t$ being an estimate of $\mu_t$, reflects the current point estimate of the expected reward of the arm $a_t$. The second term represents the confidence we have in this estimate, resembling a typical confidence interval: $s_t(a_t) \doteq \sqrt{f(x_t, a_t)^T B_{t}^{-1}f(x_{t}, a_t)}$ reflects the uncertainty, or the standard deviation, and $\alpha$ can be viewed as a generalization of the critical value. Note also that $B_t^{-1}$ and $b_t$ are analogous to the terms `$(X^TX)^{-1}$' and `$X^TY$', respectively, appearing in the ordinary least squares estimator for a standard linear regression model with $\mathbb{E}[Y|X] = X^T\mu$. If we assume a ridge penalized estimation strategy, with penalty parameter $\lambda \geq 0$, these values are recursively computed at each time $t$ taking into account previously explored arms: $B_t \doteq \lambda \mathbb{I}_{n} + \sum_{\tau=0}^{t-1}f(x_\tau, \tilde{a}_\tau)^Tf(x_\tau,  \tilde{a}_\tau)$ and $b_t \doteq \sum_{\tau=0}^{t-1}f(x_\tau,  \tilde{a}_\tau)^TY(x_\tau,  \tilde{a}_\tau)$, where $\left\{\tilde{a}_\tau \doteq \argmax_{a_\tau \in \mathcal{A}}U_\tau(a_\tau)\right\}_{\tau = 0,1,\dots,t-1}$ are the optimal arms estimated at previous times and $\mathbb{I}_{n}$ is the identity matrix of order $n$. A schematic of the LinUCB approach is provided in Supplementary Material E.

Several variations of LinUCB were proposed in the bandit literature. These include: i) the \textit{linear associative RL} strategy~\citep{auer2002using}, based on singular value decomposition rather than ridge regression; ii) generalized linear models, aiming to accommodate more complex models either for the reward~\citep{filippi2010parametric,li2017provably} or the environment~\citep{urteaga_sequential_2019}; iii) nonparametric modeling of the reward function, such as Gaussian processes~\citep{srinivas2012information}; and iv) a neural network-based feature construction which overcomes the linear reward assumption~\citep{zhou2019neural}. More recently, in addition to the (bandit) optimization goal, attention has been given to statistical objectives. To illustrate, in a similar context as ours, i.e., behavioral science, Dimakopoulou et al.~\cite{dimakopoulou19} introduced balancing methods from the causal inference literature. Specifically, to make the algorithm less prone to bias, authors proposed to weight each observation with the estimated inverse probability of a context being observed for an arm. 
This algorithm helps to reduce bias, particularly in misspecified cases, at the cost of increased variance.

Successful applications of LinUCB in mHealth can be found in \cite{paredes2014poptherapy} and \cite{forman2019can}. 
The former developed a LinUCB-based intervention recommender system for delivering stress management strategies (upon user's request in a mobile app), with the goal of maximizing stress reduction. After four weeks of study, participants who received LinUCB-based recommendations demonstrated to use more constructive coping behaviors. Similarly, in \cite{forman2019can}, a pilot study was conducted to evaluate the feasibility and acceptability of an RL-based behavioral weight loss intervention system. Participants were randomized between a nonoptimized group, an individually-optimized group (individual reward maximization), and a group-optimized (group reward maximization) group. The study showed that the LinUCB-based optimized groups have strong promise in terms of the outcome of interest, not only being feasible and acceptable for participants and coaches, but also achieving desirable results at roughly one-third the cost.

\subsubsection{Contextual MABs with LinTS exploration} \label{Sec: LinTS}
Although Thompson sampling (TS)~\citep{thompson_likelihood_1933} has been introduced more than 80 years ago, it has only recently reemerged 
as a powerful tool for online decision-making, due to its optimal empirical and theoretical properties. Under the same linear reward assumption as in LinUCB, Agrawal and Goyal~\cite{agrawal2013thompson} proposed a generalization of TS to a contextual setting. Rooted in a Bayesian framework, the idea of TS is to select arms according to their posterior probability of being optimal, i.e., by maximizing the posterior reward distribution, at each time $t$. The policy $\pi$ at each time $t$ is thus explicitly defined as:
\begin{align} \label{eq: TS_prob}
     \pi_t(a) &= \mathbb{P}\Big(Q_t^{\boldsymbol{\pi}}(x_t, a) \geq Q_t^{\boldsymbol{\pi}}(x_t, a'), \forall a' \neq a \mid \mathbf{H_t} = \mathbf{h_t} \Big) \nonumber\\
     &=\mathbb{P}\Big(\mathbb{E}[Y_{t+1} \mid X_t = x_t, A_{t} = a] \geq \mathbb{E}[Y_{t+1} \mid X_t = x_t, A_{t} = a'], \forall a' \neq a  \mid \mathbf{H_t} = \mathbf{h_t} \Big),\quad t= 0,1,\dots,  
\end{align} 
where the conditioning term $\mathbf{H_t} = \mathbf{h_t}$ reflects the posterior nature of this probability and should not be confused with the conditioning terms of the Q-function. The TS policy has been shown to be asymptotically optimal, meaning that it matches the asymptotic lower bound of the regret introduced by Lai and Robbins~\citep{lai1985asymptotically}. 

The typical way to implement TS is iterative and involves a posterior sampling procedure (see e.g.,~\cite{chapelle2011empirical}). For example, in the common case of a Gaussian reward model with variance $\nu^2$, i.e., $Y_t|\mu,f(X_t, A_t) \sim \mathcal{N}(f(X_t, A_t)^T\mu, \nu^2)$, considering a Gaussian prior for the regression coefficients vector $\mu$, e.g., $\mu \sim \mathcal{N}(\mathbf{0}_{n}, \sigma^2 \mathbb{I}_{n})$, at each time $t$, the optimal arm is the one that maximizes the posterior estimated expected reward, or $f(X_t, A_t)^T\tilde{\mu}_t$. The posterior nature is reflected in $\tilde{\mu}_t$, which represents a sample from the estimated posterior distribution, given by $\mathcal{N}(\hat{\mu}_t, \nu^2 B_t^{-1})$; here $\hat{\mu}_t \doteq B_t^{-1}b_t$ is the posterior mean, with $B_t$ and $b_t$ defined in the same way as for LinUCB. The iterative LinTS procedure is given in Supplementary Material E.

Given the history up to time $t$ and $f(X_t, A_t)$, the LinUCB allocation policy is deterministic, in the sense that the $t$-step arm or intervention $a_t$ is uniquely determined as the one that maximizes the upper confidence bound in Eq.~\eqref{eq: UCB}; all the other arms have a null probability of being assigned. In contrast, LinTS can be regarded as a randomized scheme, where each of the admissible arms has a positive probability of being assigned to an individual, independent of their history. In other words, given the history up to time $t$ and $f(X_t, A_t)$, the LinTS allocation policy, as defined in Eq.~\eqref{eq: TS_prob}, is still random. In terms of exploration, LinUCB allows exploration through the uncertainty term $s_t(a_t)$, while LinTS achieves it through the random draws from the posterior distribution, or, equivalently, through the probability $\pi_t(a_t)$ in Eq.~\eqref{eq: TS_prob}. Note that the standard deviation of LinUCB and LinTS is of the same order. In fact, in LinTS $Y_t|\mu_t,f(x_t, a_t) \sim \mathcal{N}\left(f(x_t, a_t)^T\hat{\mu}_t, \nu^2 f(x_t, a_t)^TB_t^{-1}f(x_t, a_t)\right)$, and by definition $ f(x_t, a_t)^TB_t^{-1}f(x_t, a_t) = s_t(a_t)$.

Similarly to LinUCB, many extensions have been considered. In the mHealth literature, specifically addressing complex likelihood functions, Eckles and Kaptein~\cite{eckles19} formulated a Bootstrap TS version to replace the posterior by an online bootstrap distribution of the point estimate $\hat{\mu}_t$ at each time $t$. The approach offers improved robustness to model misspecifications (due to the robustness of the bootstrap approach), and it can be easily adapted to dependent observations, a common feature of behavioral sciences. Tackling a different issue, namely sparse and noisy data, Tomkins et al.~\cite{tomkins2021intelligentpooling} introduced \textit{Intelligent Pooling}, a generalized version of LinTS with a Gaussian mixed-effects linear model for the reward. By explicitly modeling heterogeneity between individuals and within an individual over time, the method demonstrates a better promise of personalization, even in small groups of users.

\paragraph*{Action-centered TS.}
Motivated by potential nonstationarities in mHealth problems, Greenewald et al.~\cite{greenewald17} generalized the stationary linear model of LinTS to a nonstationary and nonlinear version, where the expected reward model is formalized as:
\begin{align}\label{eq: non-stat-reward}
    \mathbb{E}(Y_{t+1}|X_t = x_t, A_t = a_t) = f(x_t, a_t)^T\mu I(a_t \neq 0) + g_t(x_t),\quad t \in \mathbb{N},
\end{align} 
with $g_t(x_t)$ being the main nonstationary component that can vary based on past information, but not on current action, and $I( \cdot)$ denoting the indicator function. Compared to LinTS, here the reward is conceived as a combination of a baseline reward (associated with a ``do nothing'' or control arm, say $a_t = 0$), which is entirely determined by $g_t(x_t)$, and a treatment or action effect (associated with noncontrol arms, say $a_t \neq 0$). The latter is a linear function of the context-action feature $f(x_t, a_t) \in \mathbb{R}^{n}$ with $\mu \in \mathbb{R}^{n}$ the unknown parameters (resembling the fixed component characterizing LinTS), but consider contexts $X_t$ chosen by an adversary based on the history up to time $t$. The term adversarial in contextual MABs can refer to the context and the reward generation mechanism: when both contexts and rewards are allowed to be chosen arbitrarily by an adversary, no assumptions on the generating process are made (see also Section~\ref{sec:MAB}), and data can be non IID. 

To avoid user habituation, caused, for example, by the delivery of too many interventions, and to prevent the algorithm from converging to an ineffective deterministic policy, a stochastic chance constraint on the size of the probabilities of delivering the noncontrol arm is considered. That is, given two fixed probability thresholds $\pi_{\text{min}}$ and $\pi_{\text{max}}$, with $0 < \pi_{\text{min}} < \pi_{\text{max}} < 1$, the probability of assigning a noncontrol arm is given by:
\begin{align*} 
    \pi_t(a) 
    = \max \Big(\pi_{\text{min}}, \min (\pi_{\text{max}}, \mathbb{P}\big(f(x_t, a)^T\mu > 0 \big) \Big),\quad a \neq 0,\nonumber
\end{align*}
where $\mathbb{P}\big(f(x_t, a)^T\mu > 0 \big)$ represents the expected treatment effect of a noncontrol arm $a$, while $\mu$ reflects the parameter posterior distribution as in LinTS. The proposed strategy, named \textit{action-centered TS}, can be viewed as a hierarchical two-step procedure, where the first step involves estimating the optimal noncontrol arm, that is, the one that maximizes the expected treatment effect or reward as in classical LinTS, and the second step is to randomly select between a control and noncontrol arm $A_t \neq 0$. To allow for better comparability with LinTS, both algorithms are described with their pseudo-code in Supplementary Material E. 

The specific use of the term ``\textit{action-centered}'' reflects the estimation procedure for the unknown parameters $\mu$: Due to the arbitrarily complex baseline reward $g_t(x_t)$, the authors propose to work with the differential reward, defined as $Y_{t+1}(X_t, A_t) - Y_{t+1}(X_t, 0) = f(X_t, A_t)^T\mu I(A_t > 0) + \varepsilon_{t+1}$, which has the scope to eliminate the component $g_t(x_t)$; this allows the derivation of an unbiased estimator, and we refer to~\citep{greenewald17} for further details. The authors also showed that the action-centered TS achieves performance guarantees similar to LinTS, while allowing for nonlinearities in the baseline reward. Additional theoretical improvements are given in \cite{krishnamurthy2018semiparametric} and \cite{kim2019contextual}. Here, a relaxation of the action-independent assumption of the component $g_t(x_t)$ in Eq.~\eqref{eq: non-stat-reward} is considered, making the reward model entirely nonparametric.

From a practical viewpoint, the algorithm has been empirically evaluated in the \textit{HeartSteps} study~\citep{liao2020personalized, klasnja2015microrandomized}, an mHealth physical activity study of great interest both in biostatistics and in the RL / bandit literature. In this context, Liao at el.~\cite{liao2020personalized}, for example, incorporated into the differential reward model an `availability' variable, stating whether the user is available to receive an intervention or not.

\subsection{Insights on current methodological differences and their drivers} \label{sec: divergence_DTR_JITAI}

So far, a broad literature documented and demonstrated the premise of RL in both types of AIs. However, the practical
methodological realities of the two remain apparently disjoint or with little commonalities. Why does Q-learning, a popular algorithm for estimating DTRs, have no practical use in JITAIs, where simplified frameworks are used? Is it reasonable to adopt simplified RL formulations given the nature of mHealth applications? Can we expect Thompson sampling, largely employed in JITAIs, to dictate the next generation of DTRs? Or ultimately, should we expect a convergence, dictated, e.g., by a greater synergy between the two areas, or should we regard them as unrelated?

Although these questions were partially covered throughout the previous sections, here we offer a systematic synthesis of the main differences and their drivers. We propose a number of insights that may guide the thinking about the future of the two areas and their potential relationship (or lack thereof), convergence, or complementarity. In Section~\ref{Sec: AIs_RealLife}, case studies supporting this discussion will be illustrated. 

\paragraph*{Offline vs online (reinforcement) learning: a different primary objective.} One of the main differences between the DTR and JITAI settings is in the role optimizing schemes such as RL have from the data acquisition to the data analysis. In DTRs, RL is \textit{typically} (we will mention a few exceptions shortly) implemented in an offline manner: we assume data have already been collected, and RL serves for estimating an optimal regime from a batch of $N$ IID data trajectories. RL does not guide the treatment decisions as new data arise, and thus has no role in the data collection process. Data are typically generated based on routine clinical practice (leading to observational data such as EHRs), or according to a randomized study where interventions are assigned with, e.g., fixed and equal probability at each stage. On the contrary, in JITAIs, RL is \textit{often} the main determinant of data collection: its scope is to determine and deliver, based on accumulating data, the right interventions in real time so as to benefit the most of the study participants. Clearly, while this process will not directly affect the sample, it will affect the intervention assignment and thus the final data.

This difference is mainly driven by the typical primary objective in each AI area, especially in the case of experimental studies, where an intervention decision can be made within the study. For DTR studies, the purpose of a potential experimental trial is generally to evaluate and compare treatments; only in subsequent analyses/phases, the goal embraces identifying and eventually assigning (to a future population) an optimal estimated regime. Nonetheless, it is worth noting that data arising from experimental studies, in particular SMARTs, are still limited. In fact, due to cost and complexity in the design and implementation, SMARTs are relatively few in number. In this landscape, the availability of observational data has guided a broad literature on DTR, leaving space for offline learning only. Yet most of the literature has focused on illustrating the application of statistical methodology, rather than informing clinical practice~\citep{mahar_scoping_2021}. One of the main challenges for an adequate translation into practice is given by the difficulty in verifying the necessary conditions for causal inference, in addition to the high heterogeneity of both patient populations and treatment implementation, which requires accurate pre-analysis considerations. We refer to Section 15.3.1.1 of~\cite{deliu_dynamic_2022} for two relevant examples on constructing DTRs with observational data.

Unlike DTRs, in JITAIs, delivering optimal AIs during the course of a program or the trial in order to optimize users' experience and engagement with the mHealth device remains the primary goal. Clearly, this trend is facilitated by the use of mobile devices and by the type of intervention (more behavioral and less clinical). Nonetheless, an increasing amount of population is using mobile devices for behavioral health support outside of experimental contexts in their everyday life. This anticipates a large amount of observational-type of data for estimating JITAIs, along with several challenges for analyses, spanning from the large portion of missing data to the high heterogeneity in terms of users' behavior in using the mHealth tool (e.g., number and distance between decision points or availability to interact with the device). 

There are some exceptions, and these may indicate a potential convergence. For example, in the context of infectious diseases, the application of a fixed randomization strategy for a prolonged period is neither ethical nor feasible. To this end, the use of TS has been proposed to learn and assign an optimal treatment strategy online~\citep{laber_optimal_2018}. Notably, the MAB choice in this setting addresses some challenges that cannot be directly addressed with offline methods such as Q-learning, including: (i) scarsity of data at the onset of an epidemic, (ii) high dimensionality and scalability with respect to state and action spaces, and (iii) a long and indefinite time horizon. Similarly, there have been theoretical works (see e.g.,~\cite{cheung_sequential_2015,wang_adaptive_2022}) that tried to incorporate online adaptations within a SMART to skew the randomization probabilities toward the most promising treatments.

\paragraph*{Simplifying assumptions and domain aspects.} Delivery of JITAIs in mHealth is carried out primarily through the simplified RL framework of contextual MABs. This simplification is essentially dictated by the strong assumption that actions $A_t$ have a momentary effect on rewards $Y_{t+1}$, but do not affect the distribution of the next states $\{X_{\tau}\}_{\tau \geq t+1}$. Essentially, one sets the discount parameter $\gamma$ to $0$, and looks for the optimal in-the-moment action. Domain knowledge envisions that such an assumption is reasonable in many mHealth applications, where information such as weather, time of the day, and GPS location, among others, is momentary, as is also its effect. However, one may certainly question the validity of such a choice or whether we should use a larger $\gamma$, full- or MDP-RL, or other strategies. In clinical conditions, this is often unrealistic: the effect of a treatment may be observed at different times, and may be affected by delayed or carryover effects. In mHealth, the study of phenomena such as habituation and delayed rewards is becoming increasingly common. Incorporating these considerations, in addition to allowing for nonstationarities (as in the action-centered TS~\cite{greenewald17}), may favor the RL methods used in DTRs and advance knowledge discovery. 

\paragraph*{Learning efficiency.} Clearly, solving a full-RL or an MDP-RL problem is much more computationally demanding than solving a contextual MAB problem. The discount rate $\gamma$ is strongly related to computational expense: the larger is the $\gamma$, or the farther we look ahead, the higher is the computational burden. By choosing small values of $\gamma$, one trades off the optimality of the learned policy for computational efficiency, which is a critical aspect in high-dimensional problems. Notably, contextual data in many mHealth applications is highly private. For this reason, much of the computation has to be done locally on mobile devices, with the risk of severely impacting battery life.

\paragraph*{Inference and real-time inference.} A key aspect that has been extensively studied in DTRs is the problem of inference (see~\cite{deliu_dynamic_2022,tsiatis_dynamic_2021} for a recent overview). This aspect has been neglected in JITAIs, where the primary goal is oriented toward reward optimization, or alternatively, participants' benefit. Learning about intervention regimes and drawing generalized conclusions is often beyond the scope of their delivery. However, even when the focus is on the ongoing study itself, how can we support the development of high-quality JITAIs without adequately assessing the effectiveness of the sequence of interventions delivered by the mobile device? Clearly, compared to (mostly behavioral) JITAIs, delivering DTRs involves a higher risk, as each intervention (often a drug) can have a substantial impact on patients' lives. Thus, among other critical points, including e.g., cost, this ethical aspect has long limited the learning and delivery of DTRs online. 

We emphasize that adaptive data-collection settings driven by RL present major challenges for statistical inference due to e.g., potential strong imbalances in arms allocations and the underlying sequential nature of the data. The problem is nowadays well documented~\cite{deliu_efficient_2021, hadad_confidence_2021}, and recent solutions have borrowed tools from causal inference~\cite{zhang_statistical_2021}. 

\paragraph*{ } While this discussion is motivated by the apparent differences between the DTR and JITAI methods, we have shown that there are exceptions, and that the use of a specific RL method should be driven by the applied problem at hand and its peculiarities (population, disease or condition, underlying domain characteristics, ethical concerns). We acknowledge that computational costs (memory and time), as well as technological limitations, still play a dominant role in JITAIs; whereas in DTRs, the main drivers relate to ethical aspects and the costs associated with running high-quality intervention studies such as SMARTs.  

Going beyond methodology, we conclude this section by suggesting that, despite the fact that DTRs and JITAIs originated and developed within two different domains while following a similar–if not the same–goal, they could often have a complementary role. In fact, if construction of an optimized DTR is part of the objectives of an experimental study (even if secondary), JITAIs could be utilized to deliver behavioral interventions to enhance both adherence to treatment and engagement with the health study, all in support of high-quality data collection, with limited deviations from study protocol.

\subsection{Further considerations and future directions} \label{sec: future_durections}
We notice that our focus in this work has been mainly devoted to the development of optimized AIs and, more specifically, to the use of RL to solve this dynamic optimization problem. Several other aspects are crucial for rigorously, validly, and ethically operating in the space of AIs. 

First, an adequate framework for causal inference is necessary. In this work, we only mentioned it in passing (see Section~\ref{sec: causal_inference}) and assumed that this foundational block and the underlying assumptions (such as unconfounding) hold. Although RL practitioners often consider problems in which the data are unconfounded, healthcare practitioners often need to make inferences using offline RL from observational data, where unconfouness is a major concern. Thus, many RL methods may not be directly applicable in practice, and learning optimal DTRs / JITAIs should account for the fact that observed actions might be affected by unobserved confounders. To address this issue, a growing literature is studying confounding bias and possible corrections under a \textit{partially observed Markov decision process} (POMDP)~\citep{miao_identifying_2018,bennett_proximal_2023,uehara_future-dependent_2023}. This states that there exist unobserved (hidden or latent) state variables, say $Z_t  \in \mathcal{Z}_t$, such that $(X_t, Z_t, Y_{t})$ forms an MDP in the sense outlined in Section~\ref{sec:MDP}, with $t \in \mathbb{N}$; that is, $(X_{t+1}, Z_{t+1}, Y_{t+1})\ \indep\ \{(X_\tau, Z_\tau, Y_{\tau})\}_{\tau =  0}^{t-1} \mid (X_{t}, Z_{t}, A_{t})$, leading to a trajectory distribution:
\begin{align*}
P_{\boldsymbol{\pi}}^{\text{POMDP}} &\doteq 
p_0(x_0, z_0) \prod_{t \geq 0} \pi_{t}(a_{t} |x_t, z_t)p_{t+1}(x_{t+1}, z_{t+1}, y_{t+1}|x_t, z_t, a_{t}). 
\end{align*} 
Note that a POMDP allows some state variables $Z_t$ to remain unobserved by the agent; therefore, it is a weaker assumption compared to a Markovian structure. Furthermore, the unobserved state can be regarded as a source of unobserved confounding. 
Within the POMDP framework, and drawing upon prior research on causal identification utilizing proxy variables~\citep{miao_identifying_2018}, Bennett and colleagues~\citep{bennett_proximal_2023} introduced the so-called \textit{proximal reinforcement learning}.  This method aims to address confounding by employing two independent proxies for confounders, one assumed to be conditionally independent from treatments given confounders, and the other assumed to be independent from outcomes given treatment and confounders. This strategy conforms to the proximal causal inference literature, which accommodates unmeasured confounding within specific causal structures. Interested readers can explore further discussions in the works of~\cite{miao_identifying_2018,zivich_introducing_2023}.

Second, if on one side the increasing technological and computational sophistication has led to new biomedical data sources (e.g., data from mobile devices and EHRs) and new algorithmic solutions (e.g., DNN or RL), on the other side it has posed some unique new challenges to be inclusively addressed. We refer to~\citep{akerkar_big_2013} for a comprehensive survey. Among these, ethical and societal concerns about fairness, accountability, and transparency are becoming increasingly relevant (see, e.g., the cross-disciplinary view adopted by top ML conferences such as the \textit{ACM Conference on Fairness, Accountability, and Transparency}). In the context of ML and RL, fairness considerations are gaining a certain depth when it comes to decision-making in healthcare and medical research (see e.g.,~\cite{mitchell_algorithmic_2021,chien_multi-disciplinary_2022,chen_algorithmic_2023}). In particular, as optimal AIs are estimated by optimizing an average cumulative outcome, due to individuals' diversity in their responsiveness to treatment and adverse effects, the estimated optimal AI may be suboptimal, risky, or even detrimental to certain underrepresented or disadvantaged subpopulations. Efforts to address this issue have been presented in e.g.,~\cite{fang_fairness-oriented_2023,zhu_risk-aware_2024,li_quasi-optimal_2023} for both single-stage and dynamic treatment regimes. The typical approach is to work with a fairness or risk-aware optimization problem, where a constraint is placed on the tail performance (fairness;~\cite{fang_fairness-oriented_2023}) or on some unwanted side effect (risk;~\cite{zhu_risk-aware_2024,li_quasi-optimal_2023}). A detailed discussion on safe/risk-sensitive RL in healthcare is provided in Appendix A of~\cite{li_quasi-optimal_2023}. Let interest be in a single final outcome $\overset{\infty}{Y}$, and denote by $\mathcal{L}_r^{\boldsymbol{d}}\left(\overset{\infty}{Y}\right) \doteq \inf\{y\!: F(y) \geq r\}$ the $r$-th quantile of $\overset{\infty}{Y}$ under policy $\boldsymbol{d}$, with $F$ and $r \in (0,1)$ denoting the cumulative distribution function and a quantile level of interest, respectively. Focusing on fairness,~\cite{fang_fairness-oriented_2023} proposed looking at the following fairness-oriented optimization problem for the population mean $\mathbb{E}_{\boldsymbol{d}}\big(\overset{\infty}{Y}\big)$:
\begin{align*}
\max_{\boldsymbol{d}} \mathbb{E}_{\boldsymbol{d}}\Big(\overset{\infty}{Y}\Big),\quad \text{subject to}\quad \mathcal{L}_r^{\boldsymbol{d}}\Big(\overset{\infty}{Y}\Big)\geq q,
\end{align*}
where $q \in \mathbb{R}$ is a predefined threshold guarantee on the tail performance. Estimation of the expectation and quantile can be more or less complex depending on whether one considers a single or a sequence of decision rules, and we refer to the original work in~\citep{fang_fairness-oriented_2023} for theoretical analyses in both cases.

Future work may also complement this project by covering and integrating from other healthcare domains that use RL. A nonexhaustive list of examples is given in~\citep{yu_reinforcement_2023,deliu2021thesis}, which includes, among others, the design of \textit{adaptive clinical trials}~\citep{us_department_of_health_and_human_services_food_and_drug_administration_adaptive_2019}. In such settings, by utilizing and processing accumulated data in an online fashion, RL and MAB methods could contribute to making clinical trials more flexible, efficient, informative, and ethical~\citep{pallmann_adaptive_2018, villar_multi-armed_2015}. Fairness has also been the subject of recent debates~\citep{chien_multi-disciplinary_2022}. All of these aspects may deserve a dedicated space, and we aim to pursue this research direction as a separate piece of work in the near future.

\section{Reinforcement Learning in Real Life: Case Studies} \label{Sec: AIs_RealLife}

This section complements the methodological framework introduced so far with its real-world implementation. Guided by two case studies we conducted in the space of DTRs and mHealth, respectively, we: (i) illustrate the applicability of RL as well as the main challenges researchers face in applying these methods in practice; and (ii) provide a concrete illustration of the main divergence between RL methods in the two areas. We start with a brief introduction of the two studies, before summarizing and comparing their main characteristics side by side in Table~\ref{tab: RL_in_RL}.

\begin{table*}
\centering
\small
    \caption{Summary of two case studies, in DTRs and JITAIs in mHealth, respectively, that used RL} 
    \begin{tabu} to \textwidth { X[.9] | X[1.7] | X[1.5] }
        
        & \textbf{\textit{PROJECT QUIT – FOREVER FREE}} & \textbf{\textit{DIAMANTE}}\\
        \hline 
        Study Design & SMART & MRT\\
        \hline
        Primary objective & To find an optimal internet-based behavioral intervention for smoking cessation and relapse prevention (based on the first stage of the SMART only) & To develop and evaluate the effectiveness of a JITAI solution for enhancing physical activity, by means of an RL-based text-messaging system\\
        \hline
        Secondary objective & To find an optimal DTR (based on the entire two-stage SMART design) & To assess the effectiveness of the JITAI solution on the distal outcome (i.e., depression)\\
        \hline
        Role of RL in the design and analysis & Secondary, used \textit{offline} for secondary post-data collection analysis & Primary, used \textit{online} in the design (data-collection phase) during interim analyses\\
        \hline
        Number, frequency and distance between decision points & Two decision points at a minimum distance of 6 months: one at the first-stage entry (the 6-month-long \textit{PROJECT QUIT}) and one after completion of the first stage at the second-stage entry (the 6-month-long \textit{FOREVER FREE}) & Daily, with around 180 decision points over a 6-month-long study: intervention decisions are made at a random time interval (Factor T in Figure~\ref{fig: mrt-diamante}) and can distance from 14 hours to 22 hours. \\
        \hline
        Model choice: interventions and tailoring variables 
        & A parsimonious model with the statistically significant elements of the primary regression analyses:
        \begin{itemize}[leftmargin=*, noitemsep, topsep=0pt]
            \item stage-1 model included two intervention factors (each at two levels) and three covariates; 
            \item stage-2 model included two intervention arms, the three stage-1 covariates and an additional covariate represented by the intermediate outcome (quit status at the end of stage 1);
            \item interactions between interventions and covariates were included as well.
        \end{itemize}
        & A high-dimensional model including: 
        \begin{itemize}[leftmargin=*, noitemsep, topsep=0pt]
            \item all baseline variables shown to be relevant in the literature and other tyme-varying covariates;
            \item an action space given by the combinations of the $4\times5\times4$ factor levels;
            \item action-action and action-contextual interactions were also included. 
        \end{itemize} \\
        \hline
        Choice of the RL strategy for optimizing interventions & Offline learning based on: 
        \begin{itemize}[leftmargin=*, noitemsep, topsep=0pt]
            \item Q-learning with a linear model, chosen for its simplicity and interpretability;
            \item a \textit{soft-thresholding} estimator (within the Q-learning framework) to address the vexing problem of nonregularity
        \end{itemize}  
        & Online learning based on: 
        \begin{itemize}[leftmargin=*, noitemsep, topsep=0pt]
            \item the computationally efficient and randomized TS algorithm to mitigate bias and to enable causal inference \citep{rosenberger_randomization_2019};
            \item self-regularization (implemented within TS) to deal with the high dimensionality and avoid overfitting;
            \item an initial uniform random `burn-in' period or, more appropriately, an `internal pilot' to acquire some prior data to feed into the main algorithm and speed up learning
        \end{itemize}\\
        \hline
        Primary outcome, i.e., the reward variable directly targeted by the intervention & 
        A final distal outcome related to smoking cessation and defined as the seven-day point prevalence of smoking (i.e., whether or not the participant smoked even a single cigarette in the last seven days prior to the end of the study stages) 
        & A proximal outcome related to physical activity and defined as the steps change from one day to another, starting the steps count from the time an intervention message is sent 
        \\
        \hline
        Handling of missing data in the reward variable & 
        \begin{itemize}[leftmargin=*, noitemsep, topsep=0pt]
            \item Descriptive checks, revealing a more or less uniform dropout across the different intervention arms, and
            \item complete case analysis~\cite{chakraborty2010inference}, as well as sensitivity analysis of multiply-imputed data~\cite{chakraborty_study_2009},
        \end{itemize} 
        to avoid sub-optimal policies due to potentially different patterns across different interventions   
        & 
        \begin{itemize}[leftmargin=*, noitemsep, topsep=0pt]
            \item Online imputation with the \textit{last observation carried forward}, and
            \item multiple imputation as a sensitivity analysis,
        \end{itemize}
        to provide reliable final estimates and avoid harmful impacts (due e.g., to technical errors in collecting observations) on online decision making
        \\
        \hline
        Other study challenges & Inference, high-dimensionality, feature extraction, sample size considerations and power analysis (see also~\cite{deliu_dynamic_2022, laber_dynamic_2014}) & Inference, non-stationarity and delayed reward, model misspecification and noisy data, users' disengagement, sample size considerations and power analysis (see also~\cite{figueroa2020guidelines,liao2020personalized})\\
        \hline \hline
    \end{tabu}
    \label{tab: RL_in_RL}
\end{table*}

\paragraph{RL for DTRs: \textit{PROJECT QUIT -- FOREVER FREE}} Based on a two-stage SMART design, this study aimed to develop / compare internet-based behavioral interventions for smoking cessation and for relapse prevention. The primary objective, interesting the first stage, known as \textit{PROJECT QUIT}, only, was to find an optimal multifactor behavioral intervention to help adult smokers quit smoking (see~\cite{strecher_web-based_2008} for details). The second stage, known as \textit{FOREVER FREE}, was a follow-on study designed to: (i) help \textit{PROJECT QUIT} participants who quit smoking stay nonsmoking, and (ii) offer a second chance to those who failed to give up smoking at the previous stage. These two stages were then considered together with the goal of finding an optimal DTR over the entire SMART study period; this was a secondary objective of the study. RL was not used in the design phase; in other words, this is not an instance of \textit{online learning}. The RL-type learning happened \textit{offline} on completion of the data collection. Detailed results from this secondary analysis can be found in~\cite{chakraborty_study_2009,chakraborty2010inference}.

\paragraph{RL in mHealth JITAIs: \textit{DIAMANTE}} Based on an MRT design (illustrated earlier in Section~\ref{Sec: RLforAIs}), the primary objective of the trial was to evaluate the effectiveness of an RL-based text-messaging system for delivering JITAIs to encourage individuals to become more physically active. In this case, RL was implemented \textit{online}, with interventions continuously optimized according to users' time-varying individual data. To evaluate the optimized JITAI solution, users were assigned to different study groups (see Figure~\ref{fig: mrt-diamante}), including a static (nonoptimized) group and the experimental RL-based adaptive group. For further details, we refer to~\cite{aguilera2020mhealth,figueroa2020guidelines}.

\section{Conclusions} \label{Sec: Discussion}
In this work, under a unified framework that brings together DTRs and JITAIs in mHealth under the area of adaptive interventions, we showed how these problems can be formalized as RL problems. With a sincere hope to enhance synergy between the methodological and applied communities, we provided a comprehensive state-of-the-art survey on RL strategies for AIs, augmenting the methodological framework with real examples and challenges. Then, we discussed the main methodological divergences in the two AI domains. 

Notably, while the two areas are ideally sharing the same problem of finding optimal policies (in line with the RL framework), their priorities are not always aligned due to historical links or domain restrictions. DTRs are mainly focused on offline estimation and identification of causal nexuses, while JITAIs are mainly engaged in online regret performances, neglecting the problem of inference. Only recently, a small body of literature started to examine the possibility of inferential goals in JITAIs, questioning the validity of traditional statistics in adaptively-collected data~\citep{zhang_statistical_2021,hadad_confidence_2021,deliu_efficient_2021}. The ML community has led the way in addressing such issues, often borrowing tools from causal inference. For example, the ``stabilizing policy'' approach of Zhang et al.~\cite{zhang_statistical_2021} is analogous to the ``stabilized weights'' of the causal inference literature~\citep{robins2000marginal}. Similarly, the adaptively-weighted IPW estimator in Hadad et al.~~\cite{hadad_confidence_2021} is inspired by the IPW estimator in~\cite{robins2000marginal}. Furthermore, an increased attention is paid to real-time or online inference to evaluate the effectiveness of JITAIs online (see e.g.,~\cite{dimakopoulou19,dimakopoulou_online_2021}). 


Despite the insufficiently mature field of mHealth, with a relatively small number of methodological studies for a rigorous evaluation of RL methods for JITAIs, their popularity in real life has grown remarkably (see Supplementary Material A). In contrast, in DTRs, the use of RL has been extensively evaluated in theoretical works, but its application in the real world is still very limited.  
Most existing DTR studies use real data only as motivational or illustrative examples. The few clinical studies focus mainly on offline learning based on observational data (e.g., EHRs) and deep learning methodologies, which limits interpretability. The explanatory drivers may be related to: 1) the lack of existing guidelines for developing optimal, yet statistically valid, DTRs; 2) the clinical setting itself, characterized by high costs, ethical concerns, and inherent complexities, which makes experimentation hard; 3) the lack of definition of AI components and the RL dynamics for the specific disease. When defining the reward function, for instance, one may need to account for multiple objectives and the presence of unstructured data, among other prior knowledge. Even from an implementation perspective, while several software packages exist for DTRs, these are often suitable only under simplified settings, e.g., continuous and positive rewards. We recognize that the area of mHealth, mostly related to behavioral aspects rather than clinical, may have fewer concerns in terms of treatment costs and risks. 

To the best of our knowledge, this represents the first piece of work that bridges the domains of DTRs and JITAIs under a unique umbrella intersecting RL and AIs. Our hope is that such a unified common ground, where different methodological and applied disciplines can easily cooperate, would help unlock the potential of exploring the opportunity RL offers in AIs and benefiting from it in a statistically justifiable way. For example, by using the rich resources on inference made available by the DTR literature, the JITAI literature may extend its goal beyond within-trial optimization. Similarly, if SMARTs were to be used in practice more often, in addition to collecting high-quality experimental data, decisions could also be optimized online, benefiting trial participants as well (see e.g.,~\cite{cheung_sequential_2015, laber_optimal_2018}), as done in JITAIs.

We also hope that our contribution may incentivize greater synergy and cooperation between the statistical and ML communities to support applied domains in the conduct of high-quality real-world studies. We recognize that this cooperation is very timely to support both the development of real-world DTR studies and to assist the spread of mHealth applications with reliable and reproducible workflows.

\section*{Acknowledgements}
Authors would like to thank the two anonymous Reviewers and Eric Laber for the constructive feedback received. Nina Deliu acknowledges support from the NIHR Cambridge Biomedical Research Centre (BRC-1215-20014) and from Sapienza University (Finanziamenti di ateneo per la ricerca scientifica, Avvio alla Ricerca Tipo 1: AR11916B8913234D). Joseph J. Williams was supported by the Office of Naval Research (N00014-21-1-2576) and the Natural Sciences and Engineering Research Council of Canada (RGPIN-2019-06968). Bibhas Chakraborty would like to acknowledge support from the start-up funding from the Duke-NUS Medical School, as well as the Academic Research Fund Tier 2 grant (MOE T2EP20122-0013) from the Ministry of Education, Singapore.

\clearpage

\appendix

\section*{Supplementary Material}

\section{Google Scholar Search}\label{App: GoogleScholar}

\begin{figure}[ht]
    \centering
    \includegraphics[scale = 0.8]{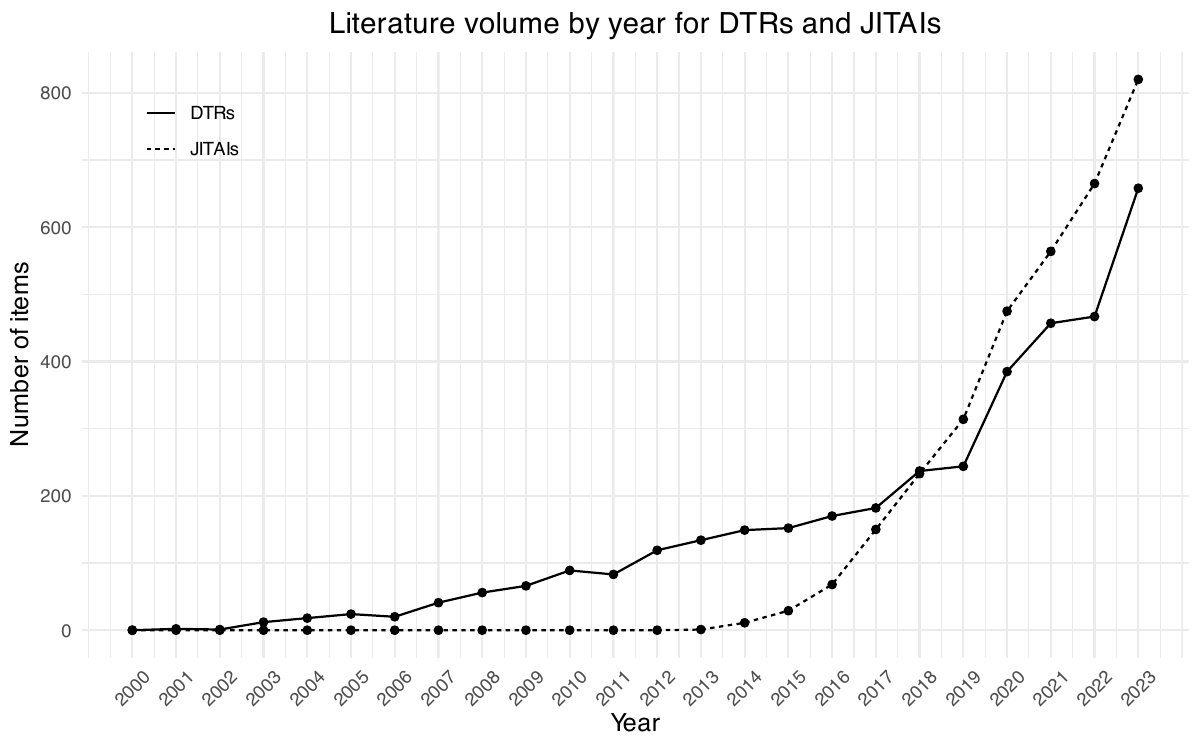}
    \caption{Literature search on Google Scholar, quantifying the interest in the literature (number of existing records) from 2000 to 2023 in the areas of DTRs and JITAIs.}
    \label{fig: GoogleScholar}
\end{figure}

The volume of literature on DTRs and JITAIs was identified on Google Scholar with the following keywords:
\begin{itemize}
    \item ``dynamic treatment regime'' OR ``dynamic treatment regimes" OR ``dynamic treatment regimen" OR ``dynamic treatment regimens'';
    \item ``just in time adaptive intervention" OR ``just in time adaptive interventions''.
\end{itemize}

Returned items contain both published articles and grey literature (e.g., preprints). Citations and patents were excluded from the literature search. A minimum screening was performed to evaluate the consistency of the identified items in relation to the searched term and the respective online publication date. Items that were incorrectly returned in correspondence with a certain date were removed from that date group.

\section{Abbreviations and Notations}

\begin{table*}[ht]
\centering
    \begin{tabu} to \textwidth { X[0.5] | X[2.1]}
        \hline
        \textbf{Symbol} & \textbf{Explanation} \\
        \hline
        \multicolumn{2}{l}{ }\\
        \multicolumn{2}{l}{Abbreviation}\\
        \hline
        AI & Adaptive intervention\\
        APP & Mobile app\\
        BMI & Body mass index\\
        C-MAB & Contextual multi-armed bandit\\
        DNN & Deep neural network\\
        DP & Dynamic programming\\
        DTR & Dynamic treatment regime / regimen\\
        EHR & electronic health records\\
        Full-RL & Full reinforcement learning\\
        IID & Independent and identically distributed\\
        IPW & Inverse probability of treatment weighting\\
        JITAI & Just-in-time adaptive intervention\\
        LinTS & Linear Thompson sampling\\
        LinUCB & Linear upper confidence bound\\
        MAB & Multi-armed bandit\\
        MDP & Markov decision process\\
        MDP-RL & RL with transition distributions following the Markov property\\
        MHealth & Mobile health\\
        ML & Machine learning\\
        MR & Meal replacement\\
        MRT & Micro-randomized trial\\
        POMDP & Partially observed Markov decision process\\
        R & Randomization\\
        RCT & Randomized-controlled trial\\
        RL & Reinforcement learning\\
        SMART & Sequential multiple assignment randomized trial\\
        TS & Thompson sampling\\
        TXT & Text message\\
        \hline
    \end{tabu}
    \label{tab: abbrev}
\end{table*}

\begin{table*}[!ht]
\centering
    \begin{tabu} to \textwidth { X[0.5] | X[2.8]}
        \hline
        \textbf{Symbol} & \textbf{Explanation} \\
        \hline
        \multicolumn{2}{l}{ }\\
        \multicolumn{2}{l}{Notation}\\
        \hline
        $i$& Trajectory, patient, user, unit, individual, subject\\
        $N$ & Number of subjects, sample size, batch\\
        $t$& Decision point, time, time point, time interval, round, stage\\
        $T$& Study horizon, last decision point\\
        $A_t\ (\mathcal{A}_t) $& Decision, action, arm, treatment, intervention, choice, option (decision space) at time $t$\\
        $X_t\ (\mathcal{X}_t)$& State, context, tailoring variables, contextual variables (state space) at time $t$\\
        $Y_t\ (\mathcal{Y}_t)$& Immediate reward, intermediate outcome, proximal outcome (outcome space) at time $t$\\
        $\overset{\infty}{Y}$& Reward, distal outcome, long-term outcome, final outcome\\
        $\mathbf{X}_t / \mathbf{A}_t / \mathbf{Y}_t$ & Set of state / action / reward variables up to time $t$ \\
        $\mathbf{H}_t\ (\mathcal{H}_t)$ & History, filtration (history space) up to time $t$, i.e., prior to time-$t$ decision\\
        $\mathbf{x}_t / \mathbf{a}_t / \mathbf{y}_t / \mathbf{h}_t$ & Observed states / actions / rewards / history up to time $t$ \\
        $\boldsymbol{d} \doteq \{d_t\}_{t\geq 0}$& Target policy of interest: set or sequence of decision rules $d_t$, (dynamic) treatment regime\\
        $\boldsymbol{\pi} \doteq \{\pi_t\}_{t\geq 0}$ & Exploration policy generating the action data; in online learning, $\boldsymbol{d} = \boldsymbol{\pi}$\\
        $P_{\boldsymbol{d}}$ & Trajectory distribution under policy $\boldsymbol{d}$\\
        $p_t$ & State-reward transition probability distribution at time $t$\\
        $\mathbf{R}_t$ & Return, i.e., cumulative sum of immediate rewards, after time-$t$ decision\\
        $\gamma$ & Discount rate specifying the current value of future rewards\\
        $V_t^{\boldsymbol{d}}$ & Value function or state-value function of policy $\boldsymbol{d}$ at time $t$\\
        $Q_t^{\boldsymbol{d}}$ & Q-function, quality function, action-value function of policy $\boldsymbol{d}$ at time $t$\\
        $\boldsymbol{\pi}^*$ / $\boldsymbol{d}^*$& Optimal policy\\
        $\hat{\boldsymbol{\pi}}^*$ / $\hat{\boldsymbol{d}}^*$& Estimated optimal estimated policy\\
        $A^*_t$ & Optimal action at time $t$\\
        $V_t^*$/$Q_t^*$ & Optimal value function / optimal Q-function at time $t$\\
        $K$ & Number of treatments, actions, arms, choices, etc\\
        $\Delta(a)$ & Immediate regret of action $a$\\
        $\text{Reg}(T)$ & Total regret over the horizon $T$\\
        $\mathcal{Q}_t$ & Approximation space for the Q-function at time $t$\\
        $\theta_t / \theta_t^*\ (\Theta_t)$ & Unknown parameter (parameter space) at time $t$\\
        $\hat{\theta}_t$ & Parameter estimate at time $t$\\
        $W, b$ ($\hat{W}, \hat{b}$) & Unknown weight and bias parameters of the neural network (their estimates)\\
        $L$ & Number of layers in a neural network\\
        $\epsilon$ & Exploration parameter of the $\epsilon$-greedy strategy\\
        $c$ & Absorbing state, representing a loss-to-follow-up, e.g., death, event\\
        $\mathcal{X}^\star$ & Augmented state $\mathcal{X}^\star \doteq \mathcal{X} \cup \{c\}$\\
        $\emptyset$ & Empty set\\
        $\mathcal{A}_x$ & An action space defined by covariate information $x$\\
        $K_x$ & A number of arms defined by covariate information $x$\\
        $\{u\}$ & Undefined set\\
        $\widetilde{T}$ & Stopping time defined by the absorbing state $c$\\
        $f / g / I $ & Context-action feature / context function / indicator function\\
        $\nabla_a$ & Gradient or derivative with respect to $a$\\
        $\hat{M} / \hat{D} / \hat{S} / \hat{\Lambda}$ & Estimating quantities for an unknown parameter\\
        $\lambda$ / $\alpha$ & Penalty parameter / tuning parameter\\
        $\mathcal{P}$ & Penalty function\\
        $\mu$ / $\hat{\mu}$ / $\tilde{\mu}$ & Unknown parameter / parameter estimate / parameter draw\\
        $U_t$ / $\hat{U}_t$ & Upper bound / estimated upper bound at time $t$\\
        $Z_t\ (\mathcal{Z}_t$) & Unobserved, hidden, or latent state (latent state space) at time $t$\\
        $\mathcal{L}_r^{\boldsymbol{d}} $ & Quantile at level $r \in (0,1)$ under policy $\boldsymbol{d}$\\
        $q$, $r$ & Predefined thresholds\\
        $\mathbf{0}_{n}$ / $\mathbb{I}_{n}$ & Zero-value vector of size $n$ / identity matrix of size $n$ \\
        $\nu^2 /  \sigma^2 / s_t$ & Variance parameters\\
        $\mathcal{N}$ & Normal distribution\\
        $\mathbb{N}$ / $\mathbb{R}$ / $\mathbb{R}_{+}$& Set of natural numbers, including $0$ / set of real numbers / set of real positive numbers\\
        $n$ & Positive constant, denoting the dimension of a space, e.g., $\mathbb{R}^n$\\
        $\mathbb{E}$ / $\mathbb{P}$ & Expectation /probability operator\\
        $\hat{\mathbb{E}}_N$ / $\hat{\mathbb{P}}_N$ & Empirical mean / probability based on a sample of size $N$\\
        $\pi_{\text{min}}$ / $\pi_{\text{max}}$ & Known probability thresholds\\
        $\varepsilon_t$ & Error or noise term at time $t$\\
        \hline
    \end{tabu}
    \label{tab: notation}
\end{table*}

\clearpage

\section{Q-learning with Function Approximation}\label{App: Q_learning_LR}

Several Q-learning function approximators have been proposed in the literature, including linear regression, decision trees, or neural networks. As Q-functions are conditional expectations, the first natural approach to model them is through linear regression models. 
Following the notation introduced in the main paper, and letting $\theta_t \doteq (\beta_t, \psi_t)$, Chakraborty and colleagues \cite{chakraborty_statistical_2013} proposed to parametrize the stage-specific optimal Q-functions as
\begin{align}\label{eq: model-q}
    Q_t^*(\mathbf{H_t}, A_t; \beta_t, \psi_t) = \beta^T_t \mathbf{H_{t0}} + (\psi^T_t \mathbf{H_{t1}})A_t,\quad t \in [0,T],
\end{align}
where $\mathbf{H_{t0}}$ and $\mathbf{H_{t1}}$ are two (possibly different) vector summaries of the history $\mathbf{H_{t}}$, with $\mathbf{H_{t0}}$ denoting the ``main effect of history'' and $\mathbf{H_{t1}}$ denoting the ``treatment effect of history''. The collections of variables $\mathbf{H_{t0}}$ are often termed \textit{predictive}, while $\mathbf{H_{t1}}$ are called \textit{prescriptive} or \textit{tailoring variables}. Parameter estimates $\hat{\theta}_t \doteq (\hat{\beta}_t, \hat{\psi}_t)$ are obtained by solving suitable estimating equations such as \textit{ordinary least squares} (OLS) or \textit{weighted least squares} (WLS). Given a sample $\left\{X_{0i}, A_{0i}, Y_{1i},\dots, X_{Ti}, A_{Ti}, Y_{(T+1)i}, X_{(T+1)i} \right\}_{i=1}^N$ of IID trajectories, the WLS estimator--whose choice might be dictated by heteroschedastic errors--derives $\hat{\theta}_t$ by solving
\begin{align*}
    0 &= \sum_{i=1}^N \frac{\partial Q_t^*(\mathbf{H_{ti}}, A_{ti}; \theta_t)}{\partial \theta_t} \Sigma_t^{-1}(\mathbf{H_{ti}}, A_{ti}) \\& \times [ Y_{(t+1)i} + \max_{a_{(t+1)i} \in \mathcal{A}_{(t+1)i}} Q_{t+1}^*(\mathbf{H_{(t+1)i}}, a_{(t+1)i}; \hat{\theta}_{t+1}) - Q_t^*(\mathbf{H_{ti}}, A_{ti}; \theta_t)],
\end{align*}
where $\Sigma_t$ is a working variance-covariance matrix. Taking $\Sigma_t$ as a constant yields the OLS estimator. 

As noticed first by~\cite{robins2004optimal} for G-estimation, and then by~\cite{chakraborty2010inference} for Q-learning, the treatment effect parameters at any stage prior to the last can be nonregular under certain longitudinal distributions of the data. Q-learning, for instance, involves modeling nonsmooth, nonmonotone functions of the data, which complicates both the regression function and the associated inference. In the specific modeling assumption of Eq. (18), due to the argmax operator involved in Q-learning, $\hat{\psi}_t$ is a \textit{nonregular} estimator, and inferential problems arise when $\hat{\psi}^T_t \mathbf{H_{t1}}$ is close to zero, leading to nondifferentiability at that point. In DTRs, this can occur, for example, when two or more treatments produce (nearly) the same mean optimal outcome. To solve this problem, adapting previous work in the context of G-estimation~\cite{moodie2010estimating}, \cite{chakraborty2010inference} proposed two alternative ways to shrink or threshold values of $\hat{\psi}^T_t \mathbf{H_{t1}}$ close to zero. In a similar spirit, \cite{song2015penalized} and \cite{goldberg2013adaptive} proposed minimizing a penalized version of the objective in the first step of Q-learning, where the penalty is given by a function $\mathcal{P}_{\lambda}(|\psi^T_t \mathbf{H_{t1}}|)$ with the tuning parameter $\lambda$, while \cite{fan2019smoothed} introduced the \textit{smoothed Q-learning} dictated by the use of a modified version of $\hat{\psi}^T_t \mathbf{H_{t1}}$, given by $\big(\hat{\psi}^T_t \mathbf{H_{t1}}\big) K_\alpha(\hat{\psi}^T_t \mathbf{H_{t1}})$. Here, $K_\alpha(x) \doteq K(x/\alpha)$, with $\alpha > 0$ a smoothing parameter and $K(\cdot)$ a kernel function that admits a probability density function. Another proposal to conduct inferences for the estimated Q-function parameters arised in \cite{chakraborty2013inference}, where a general method for bootstrapping under nonregularity, i.e., \textit{m-out-of-n bootstrap} was presented. Subsequently, \cite{laber2014interactive} derived a new \textit{interactive Q-learning} method, where the maximization step is delayed, by adding an additional step between the Eqs. (14) and (15) of the main paper. This enables all modeling to be performed before the nonsmooth and nonmonotone transformation.

\section{A-learning with Function Approximation} \label{app: A-learning}
A-learning, where `A' stands for the `advantage' incurred if the optimal treatment were given as opposed to what was actually given, is used to describe a class of alternative methods to Q-learning, predicated on the fact that it is not necessary to specify the entire Q-function to estimate an optimal regime. Models can be posited only for parts of the expectation involving \textit{contrasts} among treatments, rather than modeling the conditional expectation itself, as in Q-learning. Recalling that $\boldsymbol{d}^* \doteq \{d_t^*\}_{t=0,\dots,T}$ denotes the optimal DTR and, denoting by $\underline{\boldsymbol{d}}^*_t \doteq \{d_\tau^*\}_{\tau=t,\dots,T}$ the optimal policy from time $t$ onward, by $\boldsymbol{d}^{\text{ref}} \doteq \{d_t^{\text{ref}}\}_{t=0,\dots,T}$ a reference regime with which we want to make comparisons, and by $0$ the standard or placebo treatment, examples of contrast include:
\begin{align*}
        \mathbb{E}[Y_{t+1}^{\mathbf{a_{t-1}}, a_t, \underline{\boldsymbol{d}}^*_{t+1}} |\mathbf{H_t}=\mathbf{h_t}] &- \mathbb{E}[ Y_{t+1}^{\mathbf{a_{t-1}}, d_t^{\text{ref}}, \underline{\boldsymbol{d}}^*_{t+1}} |\mathbf{H_t}=\mathbf{h_t}],
        \\
        \mathbb{E}[Y_{t+1}^{\mathbf{a_{t-1}},a_t, \underline{\boldsymbol{d}}^*_{t+1}} |\mathbf{H_t}=\mathbf{h_t}] &- \mathbb{E}[ Y_{t+1}^{\mathbf{a_{t-1}}, 0, \underline{\boldsymbol{d}}^*_{t+1}} |\mathbf{H_t}=\mathbf{h_t}],
        \\
        \mathbb{E}[Y_{t+1}^{\mathbf{a_{t-1}}, a_t, \underline{\boldsymbol{d}}^*_{t+1}} |\mathbf{H_t}=\mathbf{h_t}] &- \mathbb{E}[ Y_{t+1}^{\mathbf{a_{t-1}}, d^*_t, \underline{\boldsymbol{d}}^*_{t+1}} |\mathbf{H_t}=\mathbf{h_t}],
\end{align*}
with 
$Y^\mathbf{a}$ the potential outcome associated with policy $\mathbf{a}$. \textit{Optimal blip-to-reference} (first expression) and \textit{optimal blip-to-zero} (second expression) evaluate the removal of a unit (`blip') of the treatment effect at stage $t$ on the subsequent mean outcome, when the optimal DTR $\boldsymbol{d}^*_{t+1}$ is followed from $t+1$ onward: the blips are represented by the reference treatment ${d}_t^{\text{ref}}$ and the 0 treatment, respectively. The last expression evaluates the increase in the expected potential outcome we forego by selecting $a_t$ rather than the optimal action $d^*_t$ at time $t$. This is called \textit{regret}, and is analogous to the notion of regret in MABs (see Section 3.3.2 of the main paper).

While \cite{robins2004optimal} advocates optimal blip functions and \cite{murphy_optimal_2003} regrets, \cite{moodie2007demystifying} demonstrated that they are mathematically equivalent. In addition, both proposed a structural nested mean model for each of the $t$ conditional intermediate causal effects, or contrasts. 
However, the two authors proposed different estimation techniques: \cite{robins2004optimal} uses backward recursive G-estimation, while~\cite{murphy_optimal_2003} uses a technique known as \textit{iterative minimization of regrets}. Thus, we distinguish the two approaches as \textit{contrast-based A-learning} and \textit{regret-based A-learning} and discuss them below. 
Comparing A-learning with Q-learning, \cite{schulte2014q} showed that Q-learning is more efficient when all models are correctly specified and the propensity model required in A-learning is misspecified. If the Q-function is misspecified, A-learning outperforms Q-learning. Finally, with both propensity and Q-learning models misspecified, there is no general trend in efficiency of estimation across parameters that might recommend one method over the other.

\begin{description}[leftmargin=0pt]
\item[Contrast-based A-learning.]
We define the \textit{optimal contrast-} or the \textit{optimal C-function} $C_t^*(\mathbf{H_t}, A_t)$ at time $t$ as the expected difference in potential outcomes when using a reference regime $d_t^{\text{ref}}$ instead of $a_t$ at time $t$, and subsequently receive the optimal regime $\underline{\boldsymbol{d}}_{t+1}^* \doteq \{ d_{\tau}^*\}_{\tau = t+1,\dots,T}$. It is basically the optimal blip-to-reference given in Section 4.1.2 (main paper) with $g$ the identity function, i.e.,
\begin{align*}
    C_t^*(\mathbf{H_t}, A_t) \doteq \mathbb{E}(Y(\mathbf{A_{t-1}}, A_t, \underline{\boldsymbol{d}}_{t+1}^*) - Y(\mathbf{A_{t-1}}, d_t^{\text{ref}}, \underline{\boldsymbol{d}}_{t+1}^*) | \mathbf{H_t}).
\end{align*} 

For simplicity, here we consider only the case of two treatment options coded as $0$ and $1$, i.e., $\mathcal{A}_t = \{0,1\}$ for all $t \in [0,T]$, and let the standard or placebo ``zero-treatment'' to be the reference treatment, i.e., $d_t^{\text{ref}} = 0$, leading to an equivalence between blip-to-reference and blip-to-zero expressions. To determine an optimal DTR, we begin by defining an approximation space for the contrast functions, e.g., $\mathcal{C}_t \doteq \left \{C_t(\mathbf{h_t}, a_t; \psi_t): \psi_t \in \Psi_t \right \}$, with $\psi \in \Psi_t$, a subset of the Euclidean space. Then, in a backward fashion, starting from $t = T$, and denoting the propensity to receive treatment $A_T = 1$ in the observed data with $\pi_T(A_T|\mathbf{h_T}) = \mathbb{P}(A_T = 1 | \mathbf{H_T} = \mathbf{h_T})$, we obtain a consistent and asymptotically normal estimator for $\psi_T$ by G-estimation~\cite{robins2004optimal}, i.e., by solving estimating equations of the form:
\begin{align}\label{eq: g-estim}
    0 = \sum_{i=1}^N \lambda_T(\mathbf{H_{Ti}}, A_{Ti})[A_{Ti} -  \pi_T(A_{Ti}|\mathbf{H_{Ti}})] [Y_{(T+1)i} -  A_{Ti}C_T^*(\mathbf{H_{ti}}, A_{Ti}; \psi_T) - \theta(\mathbf{H_{Ti}}, A_{Ti})],
\end{align}
for arbitrary functions $\lambda_T(\mathbf{H_T}, A_{Ti})$ of the same dimension as $\psi_T$ and arbitrary functions $\theta_T(\mathbf{H_T}, A_{Ti})$. To implement the estimation of $\psi_T$, one may adopt parametric models for all the unknown functions, including $\pi_T(A_{Ti}|\mathbf{H_{Ti}})$ if the randomization probabilities are not known, i.e., in observational studies. Under certain conditions, \cite{schulte2014q} report that an optimal choice for $\lambda_T(\mathbf{H_{Ti}}, A_{Ti}; \psi_T)$ is given by $\partial/\partial\psi_TC_T^*(\mathbf{H_{ti}}, A_{Ti}; \psi_T)$. Once we get estimates $\hat{\psi}_T$, the contrast-based A-learning algorithm iteratively proceeds by estimating $\hat{\psi}_{T-1},\hat{\psi}_{T-2},\dots,\hat{\psi}_{0}$. Finally, in this two-treatment setting, the optimal DTR (assuming it is unique) is given by the one that leads to a positive C-function, i.e.,
\begin{align*}
    \hat{d}_t^*(\mathbf{H_t}) = d_t^*(\mathbf{H_t}; \hat{\psi}_t) = \mathbb{I}(C_t^*(\mathbf{H_t}, A_t; \hat{\psi}_t) > 0), \quad \forall t \in [0,T].
\end{align*}
Notice that, as the additional models specified in Eq.~\eqref{eq: g-estim} are only adjuncts to estimating $\psi_T$, as long as at least one of these models is correctly specified, Eq.~\eqref{eq: g-estim} will provide a consistent estimator for $\psi_T$ (this property is called \textit{double robustness property}). In contrast, Q-learning requires correct specification of all Q-functions. An intermediate approach between G-estimation and Q-learning, which provides double-robustness to model misspecification and requires fewer computational resources compared to the former, was later introduced by~\cite{wallace2015doubly} as the \textit{dynamic weighted ordinary least squares}.
\bigskip 

\item[Regret-based A-learning.] Rather than modeling a contrast defined as the expected difference in outcome when using a reference regime $d_t^{\text{ref}}$ instead of $a_t$ at time $t$, \cite{murphy_optimal_2003} proposed to model the regret. Denoting it by $\mu_t^*$, it is defined as $\mu_t^*(\mathbf{H_t}, A_t) \doteq \mathbb{E}(Y(\mathbf{A_{t-1}}, A_t, \underline{\boldsymbol{d}}_{t+1}^*) - Y(\mathbf{A_{t-1}}, d_t^*, \underline{\boldsymbol{d}}_{t+1}^*) | \mathbf{H_t})$, for $t \in [0,T]$. Here, the advantage/regret is the gain/loss in performance obtained by following action $A_t$ at time $t$ and thereafter the optimal regime $\underline{\boldsymbol{d}}_{t+1}^*$ as compared to following the optimal regime $\underline{\boldsymbol{d}}_{t}^*$ from time $t$ on. Again, to estimate the optimal treatment regime, we model the regrets by defining an approximation space for the $t$-th advantage $\mu$-function, e.g. $\mathcal{M}_t \doteq \left \{\mu_t(\mathbf{h_t}, a_t; \psi_t): \psi_t \in \Psi_t \right \}$, with $\psi \in \Psi_t$, a subset of the Euclidean space. As with Q-learning and in contrast-based A-learning, we use approximate dynamic programming and permit the estimator to have different parameters for each time $t$. However, in this case, an estimation strategy known as \textit{iterative minimization for optimal regimes}~\cite{murphy_optimal_2003} is adopted. Specifically, \cite{murphy_optimal_2003} proposed to simultaneously estimate the regret model parameter $\psi$ plus a parameter $c$ used to improve the stability of the algorithm, searching for $(\hat{\psi}, \hat{c})$ that satisfy
\begin{align}\label{eq:IMOR}
    \sum_{t=0}^T \hat{\mathbb{P}}_N\Bigg[
    &Y + \hat{c} +  \sum_{\tau=0}^T\mu_\tau(\mathbf{H_\tau}, A_\tau; \hat{\psi})-  \sum_a\mu_t(\mathbf{H_t}, a; \hat{\psi})\pi_t(a|\mathbf{H_t}; \hat{\alpha}) \Bigg] ^2 \\ 
    &\leq 
    \sum_{t=0}^T\hat{\mathbb{P}}_N \bigg[
    Y + c + \sum_{\tau \neq t}\mu_\tau(\mathbf{H_\tau}, A_\tau; \hat{\psi}) + \mu_t(\mathbf{H_t}, A_t; \psi) \sum_a\mu_t(\mathbf{H_t}, a; \psi)\pi_t(a|\mathbf{H_t}; \alpha) \bigg] ^2,\nonumber
\end{align}
for all $\psi$ and $c$, with $\hat{\mathbb{P}}_N$ denoting the empirical mean of a sample of $N$ patients. The technique proposed to find solutions to Eq.~\eqref{eq:IMOR} is an iterative search algorithm until convergence, shown to be a special case of G-estimation under the null hypothesis of no treatment effect \cite{moodie2007demystifying}. We point to the original work of \cite{murphy_optimal_2003} and \cite{moodie2007demystifying} for readers interested in this technique and its relationship with G-estimation.

\end{description}

\section{Algorithms Pseudo-codes} \label{app: pseudo-algo}

\begin{algorithm*}
\KwIn{Time horizon $T$, sample of $N$ trajectories, approximation space for the Q-functions $\mathcal{Q}_t \doteq \left \{Q_t^{\boldsymbol{d}}(\mathbf{h_t}, a_t; \theta_t): \theta_t \in \Theta_t \right \}$, for all $t = 0,\dots,T$.}

\textbf{Initialization}: Stage-($T+1$) optimal Q-function; for convenience, is typically set to $Q_{T+1}^*(\mathbf{h_{T+1}}, a_{T+1}; \hat{\theta}_{T+1}) = \hat{\mathbb{E}}[Y_{T+1}|\mathbf{H_T} = \mathbf{h_T}, A_T = a_T] = 0$.\\

\For{$t = 0, 1, 2, \dots T$}{
\textbf{Q-function parameter estimates}: get updated estimates $\hat{\theta}_{T-t}$ backward by minimizing a loss, e.g., squared-error loss:
\begin{align} \label{eq: q-le-fa}
    \hat{\theta}_{T-t} \in \argmin_{\theta_{T-t} \in \Theta_{T-t}} \hat{\mathbb{P}}_N [
    Y_{T-t+1} & + \max_{a_{T-t+1} \in \mathcal{A}_{T-t+1}} Q_{T-t+1}^*(\mathbf{h_{T-t+1}}, a_{T-t+1}; \hat{\theta}_{T-t+1}) \nonumber \\
    & - Q_{T-t}^*(\mathbf{h_{T-t}}, a_{T-t}; \theta_{T-t})
    ]^2.
\end{align}

\textbf{Optimal policy estimate}: get the $(T-t)$-time optimal regime estimate as the one that maximizes the optimal $(T-t)$-time Q-function estimate  
\begin{align} \label{eq: q-le-fa-2}
    d_{T-t}^*(\mathbf{h_{T-t}}; \hat{\theta}_{T-t}) = \argmax_{a_{T-t} \in \mathcal{A}_{T-t}} Q_{T-t}^*(\mathbf{h_{T-t}}, a_{T-t}; \hat{\theta}_{T-t}).
\end{align}
}
\textbf{end for}
\caption{Q-learning with Function Approximation \cite{murphy2005generalization}}
\label{algo: Q-learning-FA}
\end{algorithm*}

\begin{algorithm*}
\KwIn{Pre-processing real data profiles $\{ (\mathbf{H}_{0,i}, A_{0,i}), Y(\mathbf{H}_{0,i}, A_{i})\}_{i=1,\dots,N}$; $\epsilon > 0$; $\gamma > 0$.}

\textbf{Initialization}: Experience memory $\mathcal{D}_0 = \{ (\mathbf{H}_{0,i}, A_{0,i}), Q_0^{\boldsymbol{d}}(\mathbf{H}_{0,i}, A_{0,i}; \mathbf{W}_0,\mathbf{b}_0)\}_{i=1,\dots,N}$ with $\{Q_0^{\boldsymbol{d}}(\mathbf{H}_{0,i}, A_{0,i}; \mathbf{W}_0,\mathbf{b}_0)\}_{i=1,\dots,N}$ based on random parameters $(\mathbf{W}_0, \mathbf{b}_0)$.\\
\textbf{Train a DNN} with labeled data $\mathcal{D}_0$ and get estimates $(\hat{\mathbf{W}}_0, \hat{\mathbf{b}}_0)$\\

\For{$t = 0, 1, 2, \dots T$}{
\textbf{$\epsilon$-greedy step}: select a random action $a_t$ with probability $\epsilon$;\\ otherwise $a_t = \argmax_{a \in \mathcal{A}} Q_t(\mathbf{H}_{t}, a; \hat{\mathbf{W}}_t, \hat{\mathbf{b}}_t)$;\\
Execute $a_t$ in emulator and observe state transition $X_{t+1}$ and reward $Y_{t+1}(\mathbf{H}_t, a_t)$;\\
Update the experience memory data: $\mathcal{D}_{t+1} = ( \mathcal{D}_t, Y_{t+1}, X_{t+1})$;\\
\textbf{Q-learning update}: update Q-function 
\begin{align*}
    Q_t^{\boldsymbol{d}}(\mathbf{H}_{t}, a_t; \mathbf{W}_0,\mathbf{b}_0)) = Y_{t+1}(\mathbf{H}_t, a_t) + \gamma \max_{a} Q^{\boldsymbol{d}}_{t+1}(\mathbf{H}_{t+1}, a; \mathbf{W}_0,\mathbf{b}_0))
\end{align*}

\textbf{DNN Update}: get updated estimates $(\hat{\mathbf{W}}_{t+1}, \hat{\mathbf{b}}_{t+1})$ that minimize the expected loss $\big(Q_t^{\boldsymbol{d}}(\mathbf{H}_{t}, a_t; \mathbf{W}_0,\mathbf{b}_0)) - Q_t(\mathbf{H}_{t}, a; \hat{\mathbf{W}}_t, \hat{\mathbf{b}}_t)\big)^2$ based on the Q-learning update.
}
\textbf{end for}
\caption{Deep Q-Network \cite{mnih2015human,liu2017deep}}
\label{algo:DQN}
\end{algorithm*}

\begin{algorithm*}
\KwIn{ $\alpha \in \mathbb{R}_+$, $\lambda \in \mathbb{R}_+$, $T \in \mathbb{N}$, $n \geq 1$}
\textbf{Initialization}: $B_0 = \lambda\mathbb{I}_{n}, b_0 = \mathbf{0}_{n}$\\
\For{$t = 0, 1, 2, \dots T$}{
Estimate the regression coefficient $\hat{\mu}_{t} = B_{t}^{-1}b_t$\\
\For{ $a_t \in \mathcal{A}$}{
Observe feature $f(X_t = x_t, A_t = a_t)$ and estimate the upper confidence bound $\hat{U}_t(a_t)$
$$\hat{U}_t(a_t) = f(X_t = x_t, A_t = a_t)^T \hat{\mu}_{t} + \alpha \sqrt{f(X_{t} = x_t, A_t = a_t)^T B_{t}^{-1}f(X_{t} = x_t, A_t = a_t)}$$
}
\textbf{end for}\\
Select arm $\tilde{a}_t = \argmax_{a_t \in \mathcal{A}}\hat{U}_t(a_t)$ and get the associated reward $Y_{t+1}(X_t= x_t, A_t = \tilde{a}_t)$\\
Update $B_{t}$ and $b_{t}$ according to the best arm $\tilde{a}_t$
\begin{align*}
    B_{t+1} &= B_t + f(X_t = x_t, A_t = \tilde{a}_t)^Tf(X_t = x_t, A_t = \tilde{a}_t)\\
    b_{t+1} &= b_t + f(X_t = x_t, A_t = \tilde{a}_t)^TY_{t+1}(X_t = x_t, A_t = \tilde{a}_t).
\end{align*}
}
\textbf{end for}
\caption{LinUCB \cite{li2010contextual, chu2011contextual}}
\label{algo: LinUCB}
\end{algorithm*}

\begin{algorithm*}
\KwIn{ $\sigma \in \mathbb{R}$, $\nu\in \mathbb{R}$, $T \in \mathbb{N}$, $n \geq 1$, $\lambda \in \mathbb{R}_+$}
\textbf{Initialization}: $B_0 = \lambda\mathbb{I}_{n}, b_0 = \mathbf{0}_{n}$\\
\For{$t = 0, 1, 2, \dots T$}{
Estimate the regression coefficient $\hat{\mu}_{t} = B_{t}^{-1}b_t$;\\
Get posterior samples $\tilde{\mu}_t \sim \mathcal{N}(\hat{\mu}_t, \nu^2 B_t^{-1})$\\
\For{$a_t \in \mathcal{A}$}{
Observe feature $f(X_t = x_t, A_t = a_t)$ and compute the `a-posteriori' estimated expected reward, i.e., $f(X_t = x_t, A_t = a_t)^T\tilde{\mu}_t$
}
\textbf{end for}\\
Select arm $\tilde{a}_t = \argmax_{a_t \in \mathcal{A}}f(X_t = x_t, A_t = a_t)^T\tilde{\mu}_t$ and get the associated reward $Y_{t+1}(X_t = x_t, A_t = \tilde{a}_t)$;\\
Update $B_{t+1}$ and $b_{t+1}$ according to the best arm $\tilde{a}_t$
\begin{align*}
    B_{t+1} &= B_t + f(X_t = x_t, A_t = \tilde{a}_t)^Tf(X_t = x_t, A_t = \tilde{a}_t);\\
    b_{t+1} &= b_t + f(X_t = x_t, A_t = \tilde{a}_t)^TY_{t+1}(X_t = x_t, A_t = \tilde{a}_t).
\end{align*}
}
\textbf{end for}
\caption{LinTS \cite{agrawal2013thompson}}
\label{algo:TS}
\end{algorithm*}

\begin{algorithm}
\KwIn{ $ \nu \doteq R\sqrt{9n\log(T/\delta)}$, $T \in \mathbb{N}$, $n \geq 1$, $0 < \pi_{\text{min}} < \pi_{\text{max}} < 1$}
\textbf{Initialization}: $B_0 = \mathbb{I}_{n}, b_0 = \mathbf{0}_{n}$\\
\For{$t = 0, 1, 2, \dots T$}{
Estimate the regression coefficient $\hat{\mu}_{t} = B_{t}^{-1}b_t$;\\
Get posterior samples $\tilde{\mu}_t \sim \mathcal{N}(\hat{\mu}_t, \nu^2 B_t^{-1})$\\
\For{$a_t \in \mathcal{A} \setminus 0$}{
Observe feature $f(X_t = x_t, A_t = a_t)$ and compute the `a-posteriori' estimated expected reward, i.e., $f(X_t = x_t, A_t = a_t)^T\tilde{\mu}_t$}
\textbf{end for}\\
Let $a^*_t = \argmax_{a_t \in \mathcal{A}\setminus 0}f(X_t = x_t, A_t = a_t)^T\tilde{\mu}_t$;\\
Compute the probability $\pi_t$ of taking the optimal non-zero arm
\begin{align*}
   \pi_t = \max \Big(\pi_{\text{min}}, \min (\pi_{\text{max}}, \mathbb{P}\big(f(x_t, a^*_t)^T\tilde{\mu} > 0 \big) \Big);\nonumber 
\end{align*}\\
Assign arm $\tilde{a}_t = a^*_t \neq 0$ with probability $\pi_t$, or $\tilde{a}_t = 0$ with probability $1-\pi_t$;\\
Get the associated reward $Y_{t+1}(X_t = x_t, A_t = \tilde{a}_t)$;\\
Update $B_{t+1}$ and $b_{t+1}$ according to arms $a^*_t$ and $\tilde{a}_t$
\begin{align*}
    B_{t+1} &= B_t + \pi_t (1 - \pi_t) f(X_t = x_t, A_t = a^*_t)^Tf(X_t = x_t, A_t = a^*_t);\\
    b_{t+1} &= b_t + f(X_t = x_t, A_t = a^*_t)^T(I(\tilde{a}_t > 0) - \pi_t)Y_{t+1}(X_t = x_t, A_t = \tilde{a}_t).
\end{align*}
}
\textbf{end for}
\caption{Action-Centered TS \citep{greenewald17}}
\label{algo:ACTS}
\end{algorithm}

\clearpage

\clearpage

\bibliographystyle{apalike}
\bibliography{main.bib}

\end{document}